\documentclass{article} 
\usepackage{iclr2023_conference,times}

\usepackage{amsmath,amsfonts,bm}

\def\eqref#1{equation~\ref{#1}}

\def\1{\bm{1}}

\DeclareMathAlphabet{\mathsfit}{\encodingdefault}{\sfdefault}{m}{sl}
\SetMathAlphabet{\mathsfit}{bold}{\encodingdefault}{\sfdefault}{bx}{n}

\usepackage{hyperref}
\usepackage{url}

\usepackage[utf8]{inputenc} 
\usepackage[T1]{fontenc}    
\usepackage{booktabs}       
\usepackage{amsfonts}       
\usepackage{nicefrac}       
\usepackage{microtype}      
\usepackage{xcolor}         
\usepackage{siunitx}
\usepackage{dcolumn}

\usepackage{amsmath}
\usepackage{algorithm} 
\usepackage[noend]{algorithmic}  
\usepackage[algo2e]{algorithm2e} 
\usepackage{longtable}
\usepackage{multirow}
\usepackage{tikz}
\usepackage{pgfplots}

\usepackage{graphicx}
\usepackage{subcaption}
\usepackage{paralist}
\usepackage{wrapfig}

\newtheorem{definition}{Definition}[section]

\title{Can Agents Run Relay Race with Strangers? Generalization of RL to Out-of-Distribution  Trajectories} 

\author{
\textbf{Li-Cheng Lan}\textsuperscript{1} \quad \textbf{Huan Zhang}\textsuperscript{2} \quad \textbf{Cho-Jui Hsieh}\textsuperscript{1} \vspace{5pt} \\
\textsuperscript{1}University of California, Los Angeles \quad \textsuperscript{2}Carnegie Mellon University  \vspace{3pt}\\
\texttt{lclan@cs.ucla.edu \enskip huan@huan-zhang.com \enskip chohsieh@cs.ucla.edu} \\ 
}

\iclrfinalcopy 
\begin{document}
\setlength{\textfloatsep}{8pt plus 1.0pt minus 2.0pt}
\setlength{\intextsep}{8pt plus 1.0pt minus 2.0pt}

\maketitle

\begin{abstract}

In this paper, we define, evaluate, and improve the ``relay-generalization'' performance of reinforcement learning (RL) agents on the out-of-distribution ``controllable'' states. 
Ideally, an RL agent that generally masters a task should reach its goal starting from any controllable state of the environment instead of memorizing a small set of trajectories. 
For example, a self-driving system should be able to take over the control from humans in the middle of driving and must continue to drive the car safely. 
To practically evaluate this type of generalization, we start the test agent from the middle of other independently well-trained \emph{stranger} agents' trajectories.
With extensive experimental evaluation, we show the prevalence of \emph{generalization failure} on controllable states from stranger agents. For example, in the Humanoid environment, we observed that a well-trained Proximal Policy Optimization (PPO) agent, with only 3.9\% failure rate during regular testing, failed on 81.6\% of the states generated by well-trained stranger PPO agents. To improve "relay generalization," we propose a novel method called Self-Trajectory Augmentation (STA), which will reset the environment to the agent's old states according to the Q function during training. After applying STA to the Soft Actor Critic's (SAC) training procedure, we reduced the failure rate of SAC under relay-evaluation by more than three times in most settings without impacting agent performance and increasing the needed number of environment interactions. Our code is available at https://github.com/lan-lc/STA.

\end{abstract}

\section{Introduction}

Generalization is critical for deploying reinforcement learning (RL) agents into real-world applications. A well-trained RL agent that can achieve high rewards under restricted settings may not be able to handle the enormous state space and complex environment variations in the real world. 
There are many different aspects regarding the generalization of RL agents. 

While many existing works study RL generalization 
under environment variations between training and testing~\cite{Kirk2021ASO}, 
in this paper, we study the generalization problem in a simple and \emph{fixed} environment under an often overlooked notion of generalization --- a well-generalized agent that masters a task should be able to start from any ``controllable'' state in this environment and still reach the goal. For example, a self-driving system may need to take over the control from humans (or other AIs trained for different goals such as speed, gas-efficient, or comfortable) in the middle of driving and must continue to drive the car safely. We can make little assumptions about what states the cars are at when the take-over happens, and the self-driving agent must learn to drive generally.
Although this problem may look ostensibly easy for simple MDPs (e.g., a 2-D maze), most real RL agents are trained by trajectories generated by its policy, and it is hard to guarantee the behavior of an agent on all ``controllable'' states. Roughly speaking, in the setting of robotics (e.g., Mujoco environments), we can define a state as controllable if there exists at least one policy that can lead to a high reward trajectory or reach the goal from this state. Unfortunately, most ordinary evaluation procedures of RL nowadays do not take this into consideration, and the agents are often evaluated from a fixed starting point with a very small amount of perturbation~\citep{Todorov2012MuJoCoAP, Brockman2016OpenAIG}. In fact, finding these controllable states themselves for evaluation is difficult.

\begin{figure}[t]
\minipage{0.31\textwidth}

  \includegraphics[width=\linewidth]{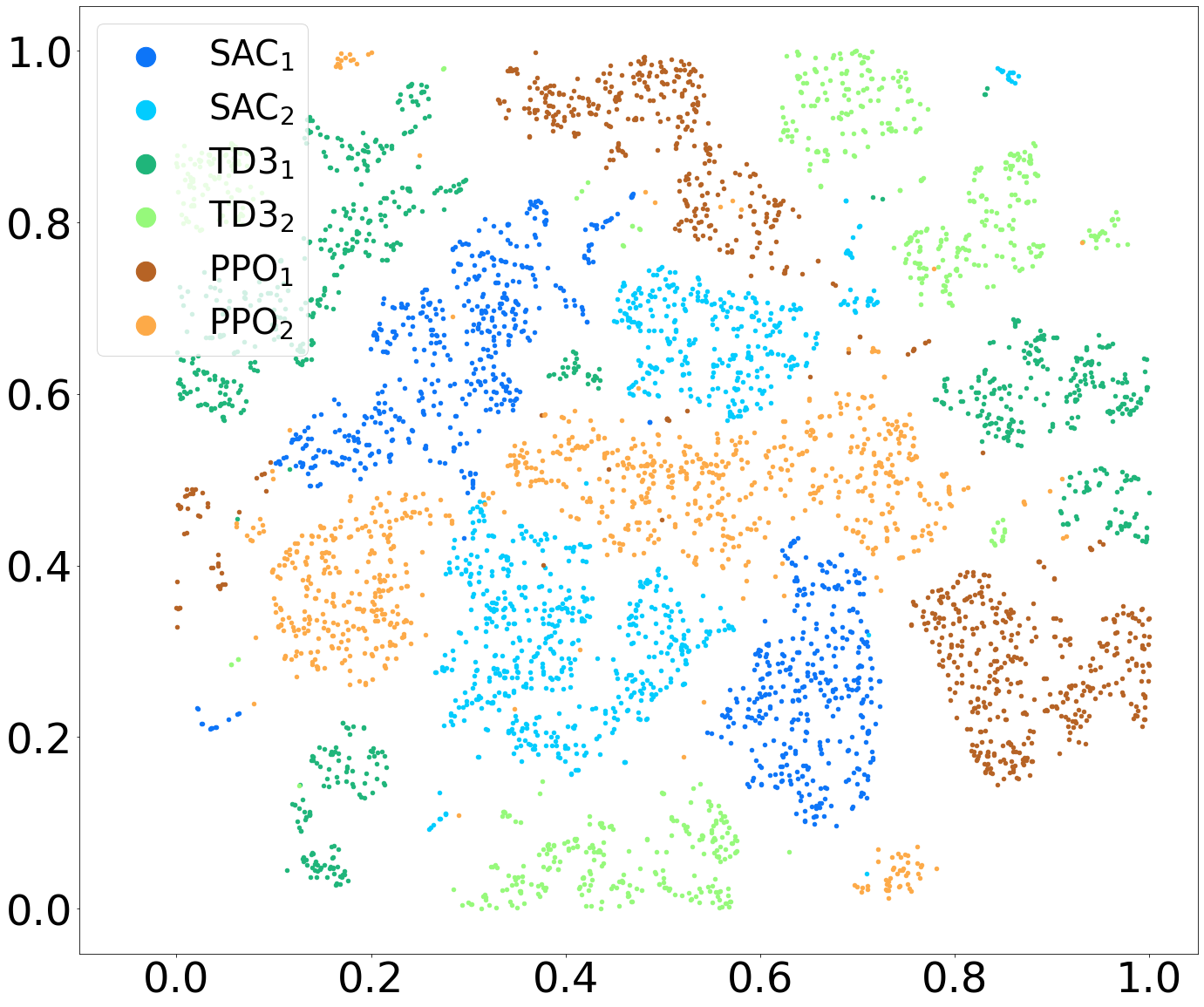}
  \subcaption{States from ordinary evaluation}
  \label{fig:distribution_1}
\endminipage\hfill
\minipage{0.31\textwidth}
  \includegraphics[width=\linewidth]{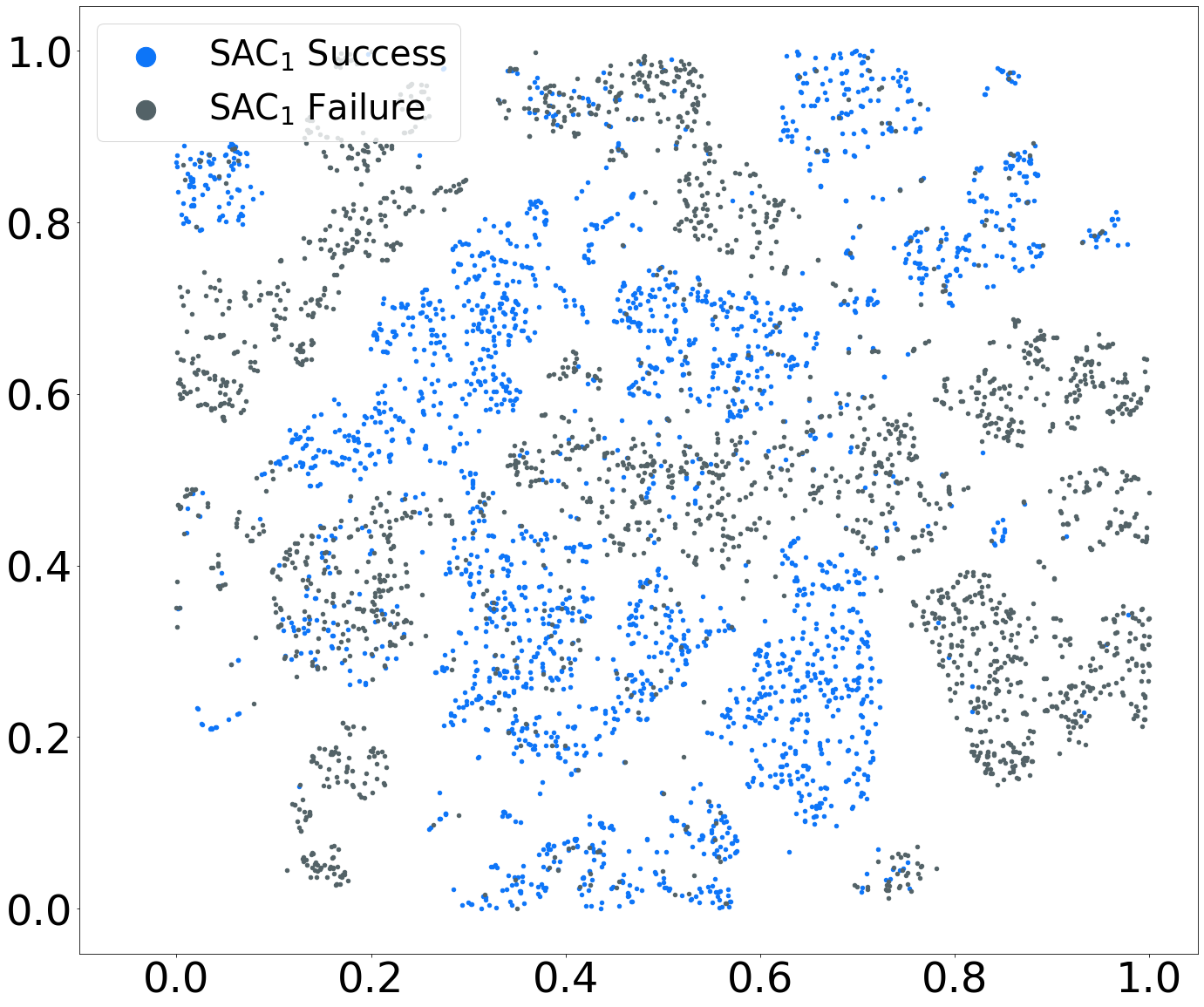}
  \subcaption{Relay-evaluation for SAC$_1$}
  \label{fig:distribution_2}
\endminipage\hfill
\minipage{0.31\textwidth}%
  \includegraphics[width=\linewidth]{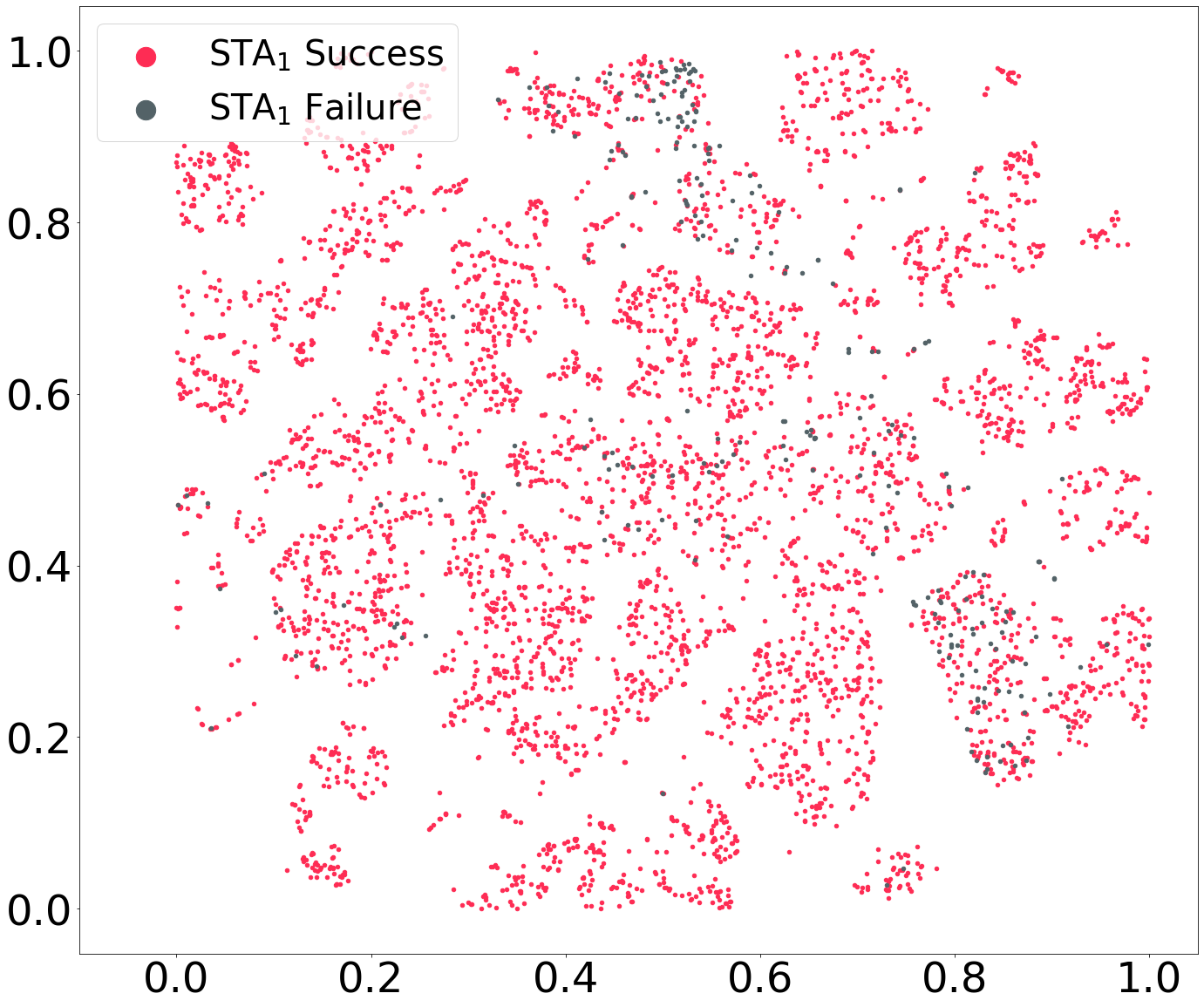}
  \subcaption{Relay-evaluation for our STA$_1$}
  \label{fig:distribution_3}
\endminipage

\caption{t-SNE of states from trajectories of 6 Humanoid agents. (a) The states of 6 agents are almost non-overlapping even for the ones trained by the same algorithm; (b) SAC$_1$ agent in (a) performs badly on controllable states from other stranger agents, indicating that it may not learn to control the robot generally; (c) our STA agent performs uniformly well when starting from the same set of states.}
\end{figure}

The \textbf{first contribution} of this work is to propose \emph{relay-evaluation}, a proxy to evaluate agent generalization performance on controllable states in a fixed environment. Relay-evaluation involves running an agent from the middle states of other independently trained agents’ high-reward trajectories. 
This is similar to running a relay race with another \emph{stranger agent}, where the stranger agent controls the robot first, and the test agent takes the reins of the robot later. It naturally finds a diverse set of controllable states for the test agents because the sampled states come from high reward trajectories of well-trained agents. The stranger agents can be trained using a variety of RL algorithms, not limited to the one used for the test agent.

Our extensive experiments on 160 agents trained in four environments using four algorithms show that many representative RL algorithms have unexpectedly high failure rates under relay-evaluation. For example, in the Humanoid environment, Proximal Policy Optimization (PPO) agents and Soft Actor Critic (SAC) agents have average failure rates of $81.6\%$ and $38.0\%$ under relay-evaluation even when the stranger agents have trained with the \emph{same} algorithms (different random seeds), which is surprisingly high compared to the original failure rates (PPO: $3.9\%$, SAC: $0.92\%$). The failure of the agents under this setting shows that they may not genuinely understand the dynamics of the environments and learn general concepts like balancing the robot to avoid failing, but rather memorize a small set of actions specifically for a limited number of states it encountered.
In Figure~\ref{fig:distribution_1}, we illustrate the t-SNE for states in trajectories of 6 agents trained with 3 algorithms and observe that even trained with the same algorithm, the states generated by different agents are quite distinctive. Figure~\ref{fig:distribution_2} shows that the SAC$_1$ agent only has a low failure rate on its own states, which is the dots colored blue in Figure~\ref{fig:distribution_1}. 

Our \textbf{second contribution} is to propose a novel training method called Self-Trajectory Augmentation (STA) that can significantly improve agents' generalization without significantly increasing training costs. We first conduct a motivative experiment to augment the initial state set of an agent during training by the states generated by a set of pretrained stranger agents, and we find that as we increase the number of stranger agents, the relay-evaluation of the agent improves significantly. However, pretraining additional models is time-consuming and may be impractical in complex environments. Therefore, we propose a novel method called Self-Trajectory Augmentation (STA), where we randomly set the agent to start from its old trajectories. Since the distribution of visited states often varies during training, reviewing an agent's old trajectories can be beneficial for generalization. After applying STA to the standard SAC training procedure, the failure rates of SAC agents are reduced by more than three times in most settings without sacrificing agent performance under ordinary evaluation, and it has minimal impact on convergence speed. In Figure~\ref{fig:distribution_3}, our STA agent is uniformly successful on most states from 6 stranger agents.

\section{Relay Evaluation}
In this section, we first introduce the definition of relay-evaluation in Sec.\ref{sec:definition}. Then we conduct extensive experiments to evaluate the relay-generalization of representative RL algorithms in Sec.\ref{sec:goat_results}.
\subsection{Notations}
\label{sec:definition}
Single-player environments of RL are normally formalized as a Markov decision process (MDP). It can be defined by a tuple $M = \langle\mathcal{S}, \mathcal{A}, \mathbb{T}, R, d_{0}\rangle$, where $\mathcal{S}$ is the state space, $\mathcal{A}$ is the action space, $\mathbb{T}(s_{t+1}|s_{t}, a_{t}): \mathcal{S} \times \mathcal{A} \times \mathcal{S} \mapsto \mathbb{R}$ is the transition function that indicates the probability of reaching state $s_{t+1}$ after playing action $a_{t}$ on state $s_{t}$; $\mathcal{R}(s_{t}, a_{t}): \mathcal{S} \times \mathcal{A} \mapsto \mathbb{R}$ is the reward function, and $d_{0}(s_0)$ is the distribution of the initial state $s_0$. With $d_0$, we can define the initial state set $\mathcal{S}_0$ as $\forall{s\in \mathcal{S}_0 \Leftrightarrow d_{0}(s) > 0}$. 
We define an agent's policy as $\pi(s)$ which simply outputs the best action.
For each $\pi(s)$, we define its the trajectories distribution as $\mathcal{T}^{\pi}$. Note that, a trajectory $\tau \sim \mathcal{T}^{\pi}$ is a list of tuple $[(s_0, a_0, r_0, s_1), (s_1, a_1, r_1, s_2), \dots, (s_{T-1}, a_{T-1}, r_{T-1}, s_{T})]$, where $s_0 \in \mathcal{S}_0$ and for each tuple $\mathbb{T}(s_{t+1}|s_{t}, a_{t}) > 0$ and $\mathcal{R}(s_{t}, a_{t}) = r_t$.  
We named the return of a trajectory $\tau$ as the sum of all the rewards of a trajectory $\sum^{T-1}_{i=0} r_i$, where $T$ is the length of the trajectory and $r_i$ is the reward of playing $a_i$ at $s_i$. 
\subsection{Problem Definition}
The goal of relay-evaluation is to evaluate test agent $\pi_\text{test}$ performance on any ``controllable'' state in the state space $\mathcal{S}$.
Here, we define ``controllable'' state as:
\begin{definition}[Controllable states]
A state $s \in \mathcal{S}$ is controllable if there exists a trajectory $\tau = \{\dots, (s_t=s, a_t, r_t, s_{t+1}), \dots\}$ where $s$ is one of its states $s_t$, and the trajectory has a high return $\sum^{T-1}_{i=0} r_i$ and $t <= T - L$. Here $T$ is the length of the trajectory, $L$ is the number of steps the agent runs after state $s$. It is used to ensure it is possible to play $L$ time steps starting from $s$ without being terminated by the environment.
\end{definition}

Particularly, we focus on the case of \emph{catastrophic failures} when starting with these controllable states. 
In our setting of continuous control environments, we define catastrophic failure as the environment being terminated by the simulator (e.g., the agent completely falls down), but one can also define catastrophic failure as getting an extremely low return. 
Conceptually, controllable states are those ``decent'' states where catastrophic failures should not happen. 
We only evaluate the test agent on controllable states since there may not exist a high return policy for a random state since some states are unrecoverable.

To find controllable states, we propose to use independently well-trained agents (which we refer to as ``stranger'' agents) that can generate high return trajectories. Therefore, in the relay-evaluation, we evaluate a test agent $\pi_\text{test}$ with the controllable states on the trajectories of another agent $\pi_\text{gen}$.
We first use $\pi_\text{gen}$ to generate $M$ trajectories and select $\eta\times M$ of them with top returns, where $\eta\in [0,1]$. 
We  then sample states $s_t, t\in \{1,2, \dots, T-L\}$ from those high return trajectories and regard those states to be controllable. 
If the test agent can continue playing for $L$ time steps without failing from a state $s_t$, then we say that the test agent passes the relay-evaluation on this state $s_t$. 
After testing $\pi_\text{test}$ on all the  controllable states generated by $\pi_\text{gen}$, we can then compute the failure rate and average return. Note that we focus on the {\bf failure rates} in most of the tables since it's surprising that most agents encounter unexpected high failure rates under relay-evaluation, while the average return is also reported in some cases to reflect the average performance. 
For a test agent, we conduct relay-evaluation on multiple generating agents, including agents independently trained by the same or different algorithms, as detailed below.

\subsection{Results of Relay-Evaluation on existing algorithms}
\label{sec:goat_results}
\label{sec:exp_setup}

\paragraph{Experiment setup.}
We conduct our experiments in four Mujoco environments in OpenAI Gym: Humanoid-v3, Walker2d-v3, Hopper-v3, and Ant-v3 with the standard setting of 1,000 steps.
We do not include HalfCheetah-v3 since the simulator does not terminate even when the robot has already fallen over, so there is no standard way to determine catastrophic failures. We first select three popular algorithms, including Soft Actor-Critic (SAC)~\citep{Haarnoja2018SoftAO}, Twin Delayed DDPG (TD3)~\citep{Fujimoto2018AddressingFA}, Proximal Policy Optimization (PPO)~\citep{Schulman2017ProximalPO}, as well as two robust training methods aiming to improve agent performance under perturbed observations, SA-PPO and ATLA-PPO~\citep{Zhang2020RobustDR, Zhang2021RobustRL}. Note that the robustness of SA-PPO and ATLA-PPO agents are not aligned with the generalization setting in our paper, but they are included to see if there exists a connection between robustness and generalization. 
All the agents have a decent average return under the ordinary evaluation which samples the initial state $s_0$ according to distribution $d_0$.
We independently train $N=10$ agents for each algorithm in each environment.
Every agent takes turns being the test agent and the stranger agent of other agents. 
For each agent $\pi_\text{gen}$, we first generate 
$M=200$ trajectories and keep the top $100$ trajectories that have a higher return. For each high-return trajectory, we sample $K=5$ states for other agents to test.

\begin{table}[t]
\caption{Failure rates (\%) of relay-evaluation using states generated by stranger agents trained with 4 algorithms, reported in the 4 rows for each environment. The ``Reference'' column shows the failure rate of the stranger agents, serving as the baseline failure rate for these controllable states. Although SAC agents achieve the lowest failure rate, they are still quite high compared to the reference in many environments. TD3 and PPO sometimes have over 90\% generalization failure rates.}
\vspace{-0.4cm}

\label{tab:main_result}
\begin{center}
\resizebox{.750\textwidth}{!}{

\begin{tabular}{ccccccc}

\multicolumn{1}{c}{\multirow{2}{*}{Environment}} &
\multicolumn{1}{c}{\multirow{2}{*}{\shortstack{Stranger\\Algorithm}}} &
\multicolumn{1}{c}{\multirow{2}{*}{Reference (\%)}} &
\multicolumn{4}{c}{Test Agent Algorithm (Failure Rate \%)} \\
\cmidrule{4-7}

& & & \multicolumn{1}{c}{SAC} & \multicolumn{1}{c}{TD3} & \multicolumn{1}{c}{PPO} & \multicolumn{1}{c}{SA/ATLA PPO} \\
\midrule
\multirow{4}*{\shortstack[c]{Humanoid}} 
& SAC & 0.92 $\pm$ 1.78    & \textbf{38.0 $\pm$ 33.9} & 83.9 $\pm$ 17.6 & 83.9 $\pm$ 16.5 & 65.1 $\pm$ 31.8 \\
& TD3 & 0.62 $\pm$ 1.34    & \textbf{33.6 $\pm$ 28.5} & 60.5 $\pm$ 30.1 & 78.4 $\pm$ 20.0 & 67.5 $\pm$ 29.9 \\
& PPO & 3.91 $\pm$ 4.91    & \textbf{48.8 $\pm$ 30.1} & 77.8 $\pm$ 24.3 & 81.6 $\pm$ 19.1 & 63.2 $\pm$ 30.9  \\
& SA-PPO & 0.12 $\pm$ 0.48 & 83.8 $\pm$ 18.0          & 96.2 $\pm$ 5.66 & 92.9 $\pm$ 11.9 & \textbf{77.0 $\pm$ 26.4} \\
\midrule[0.1pt]
\multirow{4}*{\shortstack[c]{Walker2d}} 
& SAC & 0.78 $\pm$ 2.49    & \textbf{26.9 $\pm$ 23.5} & 35.9 $\pm$ 28.5 & 87.0 $\pm$ 11.4 & 88.9 $\pm$ 11.7 \\
& TD3 & 0.31 $\pm$ 0.68    & \textbf{22.7 $\pm$ 21.7} & 36.9 $\pm$ 28.1 & 84.3 $\pm$ 14.0 & 87.0 $\pm$ 12.8 \\
& PPO & 0.00 $\pm$ 0.00    & \textbf{14.5 $\pm$ 12.9} & 25.0 $\pm$ 20.0 & 70.1 $\pm$ 21.0 & 75.1 $\pm$ 19.3 \\
& ATLA-PPO & 0.64 $\pm$ 3.20 & \textbf{19.9 $\pm$ 18.1} & 29.4 $\pm$ 24.0 & 71.1 $\pm$ 21.7 & 76.2 $\pm$ 20.0 \\
\midrule[0.1pt]
\multirow{4}*{\shortstack[c]{Hopper}} 
& SAC & 0.37 $\pm$ 0.87    & \textbf{32.6 $\pm$ 36.0} & 44.1 $\pm$ 21.1 & 62.1 $\pm$ 19.8 & 40.8 $\pm$ 30.4 \\
& TD3 & 0.53 $\pm$ 1.36    & 31.9 $\pm$ 36.5          & \textbf{19.8 $\pm$ 22.1} & 70.2 $\pm$ 17.6 & 43.8 $\pm$ 28.9 \\
& PPO & 0.85 $\pm$ 2.57    & \textbf{23.8 $\pm$ 33.5} & 34.5 $\pm$ 18.0 & 56.5 $\pm$ 20.7 & 36.8 $\pm$ 29.6 \\
& ATLA-PPO & 0.01 $\pm$ 0.02 & \textbf{30.6 $\pm$ 35.0} & 31.4 $\pm$ 20.8 & 64.1 $\pm$ 18.4 & 39.5 $\pm$ 31.6 \\
\midrule[0.1pt]
\multirow{4}*{\shortstack[c]{Ant}} 
& SAC & 0.63 $\pm$ 0.85    & \textbf{2.70 $\pm$ 1.98} & 8.98 $\pm$ 9.96 & 4.00 $\pm$ 2.68 & 3.79 $\pm$ 3.09 \\
& TD3 & 1.11 $\pm$ 1.43    & \textbf{2.90 $\pm$ 2.66} & 10.0 $\pm$ 11.6 & 3.98 $\pm$ 2.13 & 4.25 $\pm$ 3.61 \\
& PPO & 1.03 $\pm$ 1.25    & \textbf{2.62 $\pm$ 1.92} & 9.95 $\pm$ 8.44 & 3.52 $\pm$ 2.07 & 3.45 $\pm$ 3.23 \\
& ATLA-PPO & 1.26 $\pm$ 1.75 & \textbf{2.32 $\pm$ 1.67} & 7.58 $\pm$ 5.31 & 3.13 $\pm$ 1.96 & 3.76 $\pm$ 3.57 \\
\midrule
\end{tabular}
}
\end{center}
\end{table}

\paragraph{Generalization Results of Existing Algorithms.}
For each agent, we conduct relay-evaluation using the trajectories generated by all the other agents as stranger agents, trained by either the same or different algorithms.
To investigate the failure rate of test agents trained by Algorithm A when testing the trajectories of stranger agents trained by Algorithm B, we report the average failure rate for each A-B pair in Table \ref{tab:main_result}. 
Each row in Table \ref{tab:main_result} shows the results of stranger agents trained by a certain algorithm B. The ``Reference'' column shows the failure rates of the stranger agents starting from their own controllable state, which is very low. This indicates that these states are indeed controllable.
For the remaining columns, we show the failure rates of relay-evaluation, and clearly, we observe that all the agents have significantly higher failure rates than the reference failure rates. For example, in Humanoid, SAC agents have a 38\% of failure rate on other SAC agents' trajectories and even higher failure rates when the stranger agent is trained by PPO or SA-PPO.
In general, SAC agents have the lowest failure rate across all environments. This is probably due to the entropy term added to their goal function, which promotes exploration. However, according to our experiments, their state distribution still collapses into a small set of successful trajectories (Figure~\ref{fig:distribution_1}) and can not handle other agents' successful trajectories properly. SA-PPO and ATLA-PPO are robustly-trained agents that perform well under perturbations on state observations. This is reflected in its smaller failure rate on Humanoid and Hopper. However, they perform badly when taking the reins from other agents, showing relay generation is quite different from adversarial robustness since we care about all controllable states instead of slightly perturbed states. Additionally, we observe that Ant-v3 is relatively easy because the robot has four legs, which is easier to keep the robot from falling. The hardest environment is Humanoid-v3 since it has the most complicated robot to balance. The failure rate of TD3 even reaches 96.2\% on SA-PPO's trajectories. 



\paragraph{Why do agents fail?} 
To investigate why the agents perform poorly on relay evaluation, we visualize the state distribution generated by different agents. 
We select the best two Humanoid agents of SAC, TD3, and PPO with the highest average return from the same agents used in Table~\ref{tab:main_result}. We generate $200$ trajectories for each agent and sample $5,000$ states from them. We then use t-SNE \citep{van2008visualizing} to reduce the dimension of all the states into 2D and visualize them in Fig.~\ref{fig:distribution_1}. We color each dot according to which agent generate the states. For example, the best and the second best SAC agent' states are colored red and blue. We can see that the states of the six agents are well separated, which means that their distributions are very different, even when trained with the same algorithms. The results hold the same in all the environments. This observation suggests that the state distribution generated by different agents is quite different, with almost no overlap. Therefore all the other agents' trajectories are out-of-distribution for a particular agent.

\begin{wrapfigure}{r}{.4\textwidth}
\centering
\includegraphics[width=4cm]{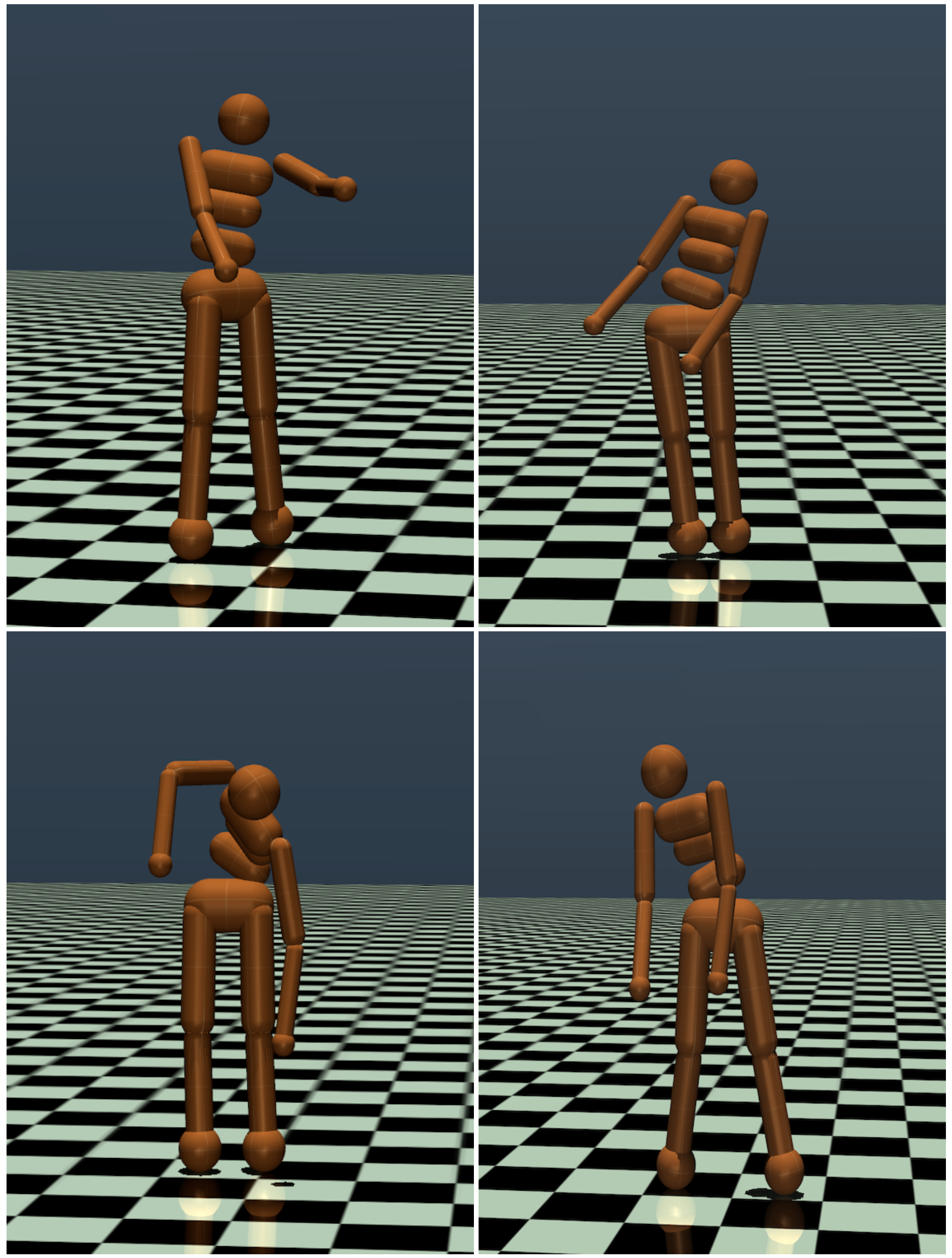}
\caption{Different SAC agents running with different postures in Humanoid.}
\label{fig:human_snapshot}
\end{wrapfigure}

\paragraph{How do the agents fail?}
We directly observe how the agents control the robot in Mujoco to understand how the agents fail and the implications of passing the relay-evaluation. Fig.~\ref{fig:human_snapshot} shows the snapshots of some different SAC agents. We observe that the agents normally maintain the balance of the upper part of the robot by fixing it to a special posture. However, different agents' postures are different, and they use different ways to keep the balance. Therefore, when an agent takes over the control from another agent, it needs to correct the posture to its own. This indicates that if an agent can pass the relay-evaluation, it generally knows how to correct the robot without losing balance, even if the starting posture is unseen.

\paragraph{Can the agents detect that it is going to fail?}
Out-of-distribution (OOD) detection has long been an important question in both RL and supervised learning, and we hope to identify the states where the agents may fail using value functions. We conduct this experiment on SAC since it has the best performance in most of the settings. Since SAC itself does not include a state value function, we use $Q(s, \pi(s))$ to serve as its evaluation of a given state, where $Q$ and $\pi$ are its action-value function and its policy. The results shown in Table~\ref{tab:detect_results} are separated into two parts. The ``Failure States'' part shows the results of states that the test SAC agents failed on, and the states that the test SAC agent passed is shown in ``Success States''. The ``Reference Return''  and ``SAC Ret'' columns show the average return of the stranger agent and the test SAC agent on the states for extra $L=500$ steps. The ``SAC Q'' shows the output $Q(s, \pi(s))$.  
We can see that the $Q(s, \pi(s))$ of the failure states is lower than the success states. 
We also notice that when the SAC agents succeed, they can perform almost as well as the stranger agents. However, the prediction $Q(s, \pi(s))$ of the success state is way lower than the prediction $Q(s, \pi(s))$  on its own states: Humanoid: $590 \pm 21$, Walker2d: $522 \pm 73$, Hopper: $363 \pm 20$, and Ant: $480 \pm 24$. This indicates that the Q function $Q(s, \pi(s))$ gives some indications about OOD and failing states. We will use this information to design our algorithms.

\sisetup{separate-uncertainty}

\begin{table}[t]
\caption{Detecting agent failure by calculating $Q(s, \pi(s))$. The numbers under the ``Failure States'' and ``Success States'' columns show the results of states that the SAC test agents have failed or succeeded on. The ``Reference Return'' (``SAC Return'') shows the average returns of the stranger agents (test agents, respectively). $Q$ is consistently lower on failure states.}
\vspace{-0.4cm}
\label{tab:detect_results}
\begin{center}
\resizebox{.95\textwidth}{!}{
\begin{tabular}{cc
S[table-format=4.0(4)]
S[table-format=4.0(4)]
S[table-format=4.0(4)]
S[table-format=4.0(4)]
S[table-format=4.0(4)]
S[table-format=4.0(4)]
}
\multicolumn{1}{c}{\multirow{2}{*}{Environment}} &
\multicolumn{1}{c}{\multirow{2}{*}{\shortstack{Stranger\\Algorithm}}} & \multicolumn{3}{c}{Failure States} & \multicolumn{3}{c}{Success States} \\
\cmidrule{3-8}
  &  & \multicolumn{1}{c}{Reference Return} & \multicolumn{1}{c}{SAC Return} & \multicolumn{1}{c}{$Q(s, \pi(s))$} & \multicolumn{1}{c}{Reference Return} & \multicolumn{1}{c}{SAC Return} & \multicolumn{1}{c}{$Q(s, \pi(s))$}  \\
\midrule
\multirow{4}*{\shortstack[c]{Humanoid}} 
& SAC & 2829 \pm 143 & 319 \pm 294 & 103 \pm 136 & 2743 \pm 96 & 2733 \pm 104 & 305 \pm 182 \\
& TD3 & 2698 \pm 114 & 388 \pm 306 & 145 \pm 144 & 2688 \pm 115 & 2746 \pm 110 & 261 \pm 166 \\
& PPO & 2939 \pm 179 & 302 \pm 233 & 81 \pm 109 & 2822 \pm 171 & 2734 \pm 110 & 270 \pm 191 \\
& SA-PPO & 3428 \pm 175 & 226 \pm 127 & 50 \pm 59 & 3288 \pm 92 & 2758 \pm 118 & 140 \pm 161 \\

\midrule[0.1pt]
\multirow{4}*{\shortstack[c]{Walker2d}} 
& SAC & 3139 \pm 292 & 386 \pm 368 & 298 \pm 135 & 2979 \pm 213 & 2834 \pm 257 & 394 \pm 112 \\
& TD3 & 3118 \pm 350 & 404 \pm 362 & 316 \pm 133 & 2974 \pm 365 & 2835 \pm 259 & 411 \pm 109 \\
& PPO & 2558 \pm 250 & 312 \pm 354 & 236 \pm 109 & 2504 \pm 247 & 2739 \pm 236 & 334 \pm 106 \\
& ATLA-PPO & 2673 \pm 375 & 275 \pm 301 & 195 \pm 117 & 2511 \pm 321 & 2721 \pm 236 & 318 \pm 114 \\
\midrule[0.1pt]
\multirow{4}*{\shortstack[c]{Hopper}} 
& SAC & 1872 \pm 66 & 1032 \pm 488 & 337 \pm 47 & 1872 \pm 62 & 1872 \pm 69 & 355 \pm 29 \\
& TD3 & 1850 \pm 95 & 801 \pm 668 & 331 \pm 55 & 1835 \pm 96 & 1873 \pm 72 & 355 \pm 34 \\
& PPO & 1819 \pm 78 & 1055 \pm 506 & 329 \pm 39 & 1808 \pm 79 & 1854 \pm 71 & 344 \pm 31 \\
& ATLA-PPO & 1889 \pm 96 & 924 \pm 558 & 330 \pm 52 & 1870 \pm 107 & 1867 \pm 72 & 352 \pm 36 \\
\midrule
\end{tabular}
}
\end{center}
\end{table}




\section{Improving the Generalization to Other Agents Trajectories}
In this section, we aim to improve the relay-generalization of the existing algorithms. We first investigate a naive method in Section~\ref{sec:algo_other}. Next, based on the investigation, we proposed our Self-Trajectory Augmentation (STA) in Section~\ref{sec:algo_self}. Finally, we conduct the experiments to evaluate both methods in Section~\ref{sec:algo_experiment}.

\subsection{A Naive Method: Leveraging Pretrained Agents' Trajectories}
\label{sec:algo_other}

To improve relay-generalization, a naive way is to let the agent ``see'' at least some of the trajectories generated by other agents during training.
One straightforward solution is to add the other agents' tuples $(s_t, a_t, r_t, s_{t+1})$ into the replay buffer as training data. However, most RL algorithms are not able to handle out-of-distribution actions except offline RL~\citep{Levine2020OfflineRL}. 
Therefore, our naive method lets the agent ``see'' other agents' trajectories by letting the agent have chances to start playing from the controllable states of some pretrained agents during training. Before training, we first pretrain additional $N_a$ agents. Next, we use states of those agents' top $\eta_{\text{naive}}=0.75$ trajectories as controllable states $\mathcal{S}^{\eta_{\text{naive}}}_{\text{pretrain}}$. During training, whenever the agent starts to generate a new trajectory, besides generating a new trajectory normally, it has $p_0$ probability to uniformly sample a state $s_{\text{pretrain}}$ from $\mathcal{S}^{\eta_{\text{naive}}}_{\text{pretrain}}$ as the initial state of this new trajectory. Starting from $s_{\text{pretrain}}$, once the agent successfully plays $L_r$ steps, we will consider it knows how to play from $s_{\text{pretrain}}$ and let the agent start a new trajectory. Our naive method is similar to training in an augmented environment that has a more diverse initial set state. Therefore, we do not need to modify the training algorithm.

\subsection{Self-Trajectory Augmentation (STA)}
\label{sec:algo_self}
\begin{wrapfigure}{R}{0.4\textwidth}
\begin{minipage}{0.4\textwidth}
    \input{algo/insert.tex}
\end{minipage}
\end{wrapfigure}
Although the naive method can improve the generalization of the agents (Section~\ref{sec:algo_experiment}), training additional agents is too time-consuming. Hence, we propose Self-Trajectory Augmentation (STA), which only uses the states of the agent's own historical trajectories to augment  the initial set. 
The framework STA algorithm is shown in Alg.~\ref{alg:select_initial_state}. For the variables, $Q, \pi$ are the agent's action value and policy that need to be trained; $\mathcal{D}$ is the replay buffer to store the training data; $\mathcal{S}_{\text{STA}}$ is the set that stores the historical states during training; $d_0$ is the original initial state distribution.
At each iteration, with probability $1-p_0$ 
 (line 5-7) we generate the trajectory as usual, which means sampling $s_0$ from $d_0$ and letting the agent plays for $L_{\max}$ steps (e.g., in Mujoco, $L_{\max}=1,000$). Otherwise, with probability $p_0$
%
%
(line 9-10), we select a state form $\mathcal{S}_{\text{STA}}$ and let the agent plays for $L_{r}$ steps. 
Since the failures  usually happen early after taking over from another trajectory, we found a small $L_r \ll L_{\max}$ is sufficient for improving relay generalization. 
After generating the trajectory, we will store the trajectory into the replay buffer (line 12) and store the qualified state (line 13) into $\mathcal{S}_{\text{STA}}$ (line 14).

\paragraph{Definition of qualified state (Alg.~\ref{alg:select_initial_state} line 13)}
Ideally, 
a state $s$ is qualified if it is controllable. However, during training, we cannot simply use the return of the trajectory to define if
its middle states are controllable. 
This is mainly due to the exploration during training -- 
the agent may play pretty well at the beginning but ends up having a low return because
the algorithm tries to explore a bad action.  
In this case, we still want to include the beginning states into $\mathcal{S}_{\text{STA}}$. Hence, we use the sum of the next $\lambda=50$ rewards as the scoring function $\text{score}(s_t) = \sum_{i=t}^{t+\lambda-1} r_i$ to measure whether a state is controllable. 
If a state's score is among the top $\eta_{\text{STA}}$ ratio of the states generated in the latest epoch $\mathcal{S}_{\text{latest}}$, the state will be added into $\mathcal{S}_\text{STA}$. 
For convenience, we maintain a threshold $\omega$ so that $\eta_{\text{STA}}$ ratio of the states in $\mathcal{S}_{\text{latest}}$ has a higher score than $\omega$. Finally, we define the function is\_qualified($s$) in line 13 as if $\text{score}(s) > \omega$.
Note that if the rewards of the environment are sparse, one might consider other scoring functions.

\paragraph{Select a state from $\mathcal{S}_{\text{STA}}$ (Alg.~\ref{alg:select_initial_state} line 9).} We select a state from $\mathcal{S}_{\text{STA}}$ with two steps. First, we random sample $N_c$ states that $\text{score}(s)\times \gamma \geq \omega_{\max}$ form 
$\mathcal{S}_{\text{STA}}$ as candidates, where $\omega_{\max}$ is the largest $\omega$ encountered during training. Although we only add states with $\text{score}(s) > \omega$ into $\mathcal{S}_{\text{STA}}$, as the agent becomes stronger, $\omega$ will also increase. Some of the states in $\mathcal{S}_{\text{STA}}$ will be considered uncontrollable since their score are too low. Hence, we use a ratio $\gamma >= 1$ to avoid low-score states ($<\omega_{\max}/\gamma$) being our candidates. For the second step, among the candidates, we want to select a state with which the current agent is unfamiliar. 
Based on the results in Table~\ref{tab:detect_results}, we use $Q(s, \pi(s))$ to predict which state is harder for the current agent. That is, we select the state with the lowest $Q(s, \pi(s))$ from the candidates as the new initial state (line 9).

\subsection{Experimental Results}
\label{sec:algo_experiment}
In this section, we evaluate the agents trained by our methods under the same settings and strange agents we used in Table~\ref{tab:main_result}. Since SAC performs best in Table~\ref{tab:main_result}, we only apply our methods based on it. In the following paragraphs, we first show that the naive method can improve the relay-generalization of SAC. Next, we show that our STA can also improve the SAC agent without slowing the training process. Moreover, STA is two times better than the naive method on the hardest controllable states.

\setlength{\textfloatsep}{10pt}
\setlength{\floatsep}{10pt}
\begin{table}[t]
\caption{The failure rates (\%) of the naive baseline with different numbers $N_a$ of pretrained agents. Training with the states from more pre-trained agents lead to better relay-evaluation performance.}
\label{tab:gsac_gen_agent_num}
\vspace{-0.4cm}

\begin{center}
\resizebox{.8\textwidth}{!}{

\begin{tabular}{cccccccc}
\multicolumn{1}{c}{\multirow{2}{*}{Environment}} &
\multicolumn{1}{c}{\multirow{2}{*}{\shortstack{Stranger\\Algorithm}}}  & \multicolumn{5}{c}{Test Algorithm (Failure Rate \%)} \\
\cmidrule{3-8}
 &  & \multicolumn{1}{c}{SAC} & \multicolumn{1}{c}{$\text{Naive}_{N_a=1}$} & \multicolumn{1}{c}{$\text{Naive}_{N_a=2}$} & \multicolumn{1}{c}{$\text{Naive}_{N_a=4}$} & \multicolumn{1}{c}{$\text{Naive}_{N_a=8}$} & \multicolumn{1}{c}{$\text{Naive}_{N_a=16}$}\\
\midrule
\multirow{4}*{\shortstack[c]{Humanoid}} 
& SAC    & 38.0 $\pm$ 33.9 & 55.7 $\pm$ 16.9 & 41.5 $\pm$ 19.7 & 24.0 $\pm$ 14.7 & \textbf{11.4 $\pm$ 7.21} & 13.2 $\pm$ 6.84   \\
& TD3    & 33.6 $\pm$ 28.5 & 40.0 $\pm$ 13.9 & 22.9 $\pm$ 11.5 & 15.0 $\pm$ 7.96 & \textbf{7.29 $\pm$ 2.79} & 7.81 $\pm$ 4.76  \\
& PPO    & 48.8 $\pm$ 30.1 & 59.3 $\pm$ 13.9 & 41.5 $\pm$ 12.5 & 30.6 $\pm$ 12.0 & \textbf{24.8 $\pm$ 3.89} & 25.0 $\pm$ 3.36  \\
& SA-PPO & 83.8 $\pm$ 18.0 & 90.8 $\pm$ 4.31 & 78.8 $\pm$ 5.83 & 71.9 $\pm$ 11.8 & 59.9 $\pm$ 9.60 & \textbf{58.1 $\pm$ 8.32} \\
\midrule[0.1pt]
\multirow{4}*{\shortstack[c]{Walker2d}} 
& SAC    & 26.9 $\pm$ 23.5 & 28.6 $\pm$ 6.53 & 20.0 $\pm$ 14.6 & 8.95 $\pm$ 5.99 & 8.14 $\pm$ 5.89  & \textbf{6.10 $\pm$ 2.34} \\
& TD3    & 22.7 $\pm$ 21.7 & 21.0 $\pm$ 6.46 & 17.2 $\pm$ 11.3 & 7.71 $\pm$ 4.31 & 4.71 $\pm$ 3.64  & \textbf{3.43 $\pm$ 1.52} \\
& PPO    & 14.5 $\pm$ 12.9 & 14.0 $\pm$ 5.83 & 10.1 $\pm$ 10.7 & 3.48 $\pm$ 1.66 & 3.05 $\pm$ 2.14  & \textbf{1.95 $\pm$ 1.35} \\
& ATLA-PPO & 19.9 $\pm$ 18.1 & 19.3 $\pm$ 4.70 & 12.0 $\pm$ 9.00 & 5.46 $\pm$ 3.59 & 3.83 $\pm$ 2.24  & \textbf{3.36 $\pm$ 1.94} \\
\midrule[0.1pt]
\multirow{4}*{\shortstack[c]{Hopper}} 
& SAC    & 32.6 $\pm$ 36.0 & 19.8 $\pm$ 15.34 & 10.2 $\pm$ 6.23 & 8.64 $\pm$ 5.50 & \textbf{4.01 $\pm$ 3.76}  &  4.35 $\pm$ 4.73  \\
& TD3    & 31.9 $\pm$ 36.5 & 19.1 $\pm$ 19.51 & 5.10 $\pm$ 4.26 & 11.5 $\pm$ 7.75 & 5.00 $\pm$ 4.51  &  \textbf{3.76 $\pm$ 5.74}  \\
& PPO    & 23.8 $\pm$ 33.5 & 17.3 $\pm$ 15.77 & 8.57 $\pm$ 7.30 & 6.38 $\pm$ 4.13 & \textbf{2.81 $\pm$ 2.93}  &  3.14 $\pm$ 4.57 \\
& ATLA-PPO & 30.6 $\pm$ 35.0 & 22.6 $\pm$ 22.65 & 5.31 $\pm$ 3.47 & 9.66 $\pm$ 7.71 & 6.46 $\pm$ 5.77  &  \textbf{3.67 $\pm$ 3.31} \\
\midrule
\end{tabular}
}
\end{center}
\end{table}

\paragraph{The naive method.} In this experiment, we evaluate whether using pretrained models can improve the relay-generalization of an agent. For the experimental settings: we add the controllable states in the top $\eta_{\text{naive}}=0.75$ ratio of each pretrain agent's trajectories into the starting set; 
we use $L_r=100$ and $p_0 \approx 0.9$ to increase the sampling frequency of the starting set. The results of using different numbers $N_a$ of pretrained agents are shown in Table~\ref{tab:gsac_gen_agent_num}. Although the pretrained agent is trained by SAC, we observe that the failure rates on all kinds of algorithms decrease as more pretrained agents are included in $\mathcal{S}^{\eta_{\text{naive}}}_{\text{pretrain}}$. 
 This indicates that the method can successfully improve relay generalization.
However, in Humanoid, the performance has converged when $N_a=8$. We think it is because Humanoid is so complex that adding more agents still cannot cover all the playing styles. Hence, the failure rates do not keep dropping after $N_a\geq 8$.
We also notice that when only using 1 pretrained agent ($N_a = 1$), the failure rate surprisingly increased in the Humanoid environment; this might be because the agent will overfit the only pretrain agent's trajectories. Additional results like using different $L_r=\{50, 100, 200, 500\}$ are shown in the Appendix. 


\sisetup{separate-uncertainty} 
\begin{table}[t]
\caption{The returns and the failure rates of SAC, our naive baseline, and STA under relay-evaluation. 
Since all the agents are trained with three million environment interactions, training an STA agent is 15.9 times faster than a $\text{Naive}_{N_a=16}$ agent. 
The ``ordinary'' rows show that STA will not decrease the regular performance. 
The rows except ``ordinary'' show that STA can achieve significantly lower failure rates under relay-evaluation compared to SAC and achieve performance similar to Naive$_{N_a=16}$.}
\vspace{-0.4cm}

\label{tab:gsac2_compare}
\begin{center}
\resizebox{.995\textwidth}{!}{

\begin{tabular}{
c
c
S[table-format=4.0(4)]
c
S[table-format=4.0(4)]
c
S[table-format=4.0(4)]
c
}
\multicolumn{1}{c}{\multirow{2}{*}{Environment}} &
\multicolumn{1}{c}{\multirow{2}{*}{\shortstack{Stranger\\Algorithm}}} & \multicolumn{2}{c}{SAC} & \multicolumn{2}{c}{$\text{Naive}_{N_a=16}$ (ours)} & \multicolumn{2}{c}{STA (ours)}\\
\cmidrule{3-8}
  &   & \multicolumn{1}{c}{Return} & \multicolumn{1}{c}{Failure Rate} & \multicolumn{1}{c}{Return} & \multicolumn{1}{c}{Failure Rate} & \multicolumn{1}{c}{Return} & \multicolumn{1}{c}{Failure Rate}\\
\midrule
\multirow{4}*{\shortstack[c]{Humanoid}} 
& ordinary  & 5650 \pm 238  & 0.62  $\pm$ 0.79 \% & 5695 \pm 159  & 2.65  $\pm$ 2.52  \% & \textbf{5899 $\pm$ 346}  & \textbf{0.45  $\pm$ 0.61} \%  \\
& SAC    & 2004 \pm 1193 & 38.0  $\pm$ 33.9 \% & 2591 \pm 841 & 13.2  $\pm$ 33.8 \% & 2815 \pm 622 &       \textbf{5.77   $\pm$ 23.3} \% \\
& TD3    & 2242 \pm 1071 & 33.6  $\pm$ 28.5 \% & 2731 \pm 621 & 7.81   $\pm$ 26.8 \% & 2862 \pm 504 &      \textbf{3.76   $\pm$ 19.0} \% \\
& PPO    & 1675 \pm 1266 & 48.8  $\pm$ 30.1 \% & 2266 \pm 1120 & 25.0 $\pm$ 43.3 \% & 2692 \pm 793 &       \textbf{9.95   $\pm$ 29.9} \%  \\
& SA-PPO & 671 \pm 996   & 83.8 $\pm$ 18.0 \% & 1382 \pm 1343 & 58.1 $\pm$ 49.3 \% & 2312 \pm 1198 &       \textbf{25.2   $\pm$ 43.4} \%\\
\midrule[0.1pt]
\multirow{4}*{\shortstack[c]{Walker2d}} 
& ordinary  & 5717 \pm 445 & 0.16  $\pm$ 0.52 \%  & \textbf{5986 $\pm$ 77} & \textbf{0.00  $\pm$ 0.00}  \% & 5736  \pm 422 & \textbf{0.00 $\pm$ 0.00} \%  \\
& SAC    & 2286 \pm 1138 & 26.9 $\pm$ 23.5 \% & 2931 \pm 658 & \textbf{5.38 $\pm$ 22.5} \% & 2784 \pm 725       & 7.33 $\pm$ 26.0 \% \\
& TD3    & 2326 \pm 1078 & 22.7 $\pm$ 21.7 \% & 2965 \pm 537 & \textbf{3.71 $\pm$ 18.9} \% & 2822 \pm 632       & 5.48 $\pm$ 22.7 \% \\
& PPO    & 2464 \pm 884  & 14.5 $\pm$ 12.9 \% & 2931 \pm 336 & \textbf{1.19 $\pm$ 10.8} \% & 2798 \pm 473       & 2.57 $\pm$ 15.8 \%\\
& ATLA-PPO & 2314 \pm 1007 & 19.9 $\pm$ 18.1 \% & 2888 \pm 449 & \textbf{2.33 $\pm$ 15.0} \% & 2782 \pm 488     & 2.83 $\pm$ 16.5 \% \\
\midrule[0.1pt]
\multirow{4}*{\shortstack[c]{Hopper}} 
& ordinary  & 3662 \pm 137 & 5.75 $\pm$ 8.85 \% & 3609 \pm 60 & \textbf{0.34  $\pm$ 0.89} \% & \textbf{3686 $\pm$ 58} & 0.42   $\pm$ 0.77  \% \\
& SAC    & 1777 \pm 390 & 32.6 $\pm$ 36.0 \% & 1846 \pm 148 & \textbf{2.59 $\pm$ 15.8} \% & 1818 \pm 323   & 9.80  $\pm$ 29.7 \% \\
& TD3    & 1654 \pm 594 & 31.9 $\pm$ 36.5 \% & 1838 \pm 265 & \textbf{3.05 $\pm$ 17.1} \% & 1790 \pm 428   & 13.8 $\pm$ 34.5 \% \\
& PPO    & 1790 \pm 337 & 23.8 $\pm$ 33.5 \% & 1832 \pm 140 & \textbf{2.62 $\pm$ 15.9} \% & 1829 \pm 245   & 8.52  $\pm$ 27.9 \% \\
& ATLA-PPO & 1711 \pm 491 & 30.6 $\pm$ 35.0 \% & 1824 \pm 257 & \textbf{4.69 $\pm$ 21.1} \% & 1841 \pm 276 & 10.2 $\pm$ 30.3 \% \\
\midrule
\end{tabular}
}
\end{center}
\end{table}

\paragraph{Self-Trajectory Augmentation (STA).} In this experiment, we compare the base SAC, our naive baseline with 16 pretrained agents, and our STA method with tuned parameters (see Appendix). Note that the environment step number of training each agent is all three million. Therefore, the training time of STA will be almost 17 times faster than the naive method. 

The comparison of both the average return and failure rate are shown in Table~\ref{tab:gsac2_compare}. We first show the performance under ordinary evaluation ($s_0\sim d_0$) of each environment (``ordinary'' rows). 
The results show that both the naive method and STA have almost equal or even higher ordinary performances, although agents are trained under augmented MDP, which is different from the original MDP.  
Moreover, the failure rates of STA in the ordinary setting are also lower than SAC in all the environments, which means that the STA agents are more robust. 
Next, we show the relay-evaluation results on different stranger algorithms. According to the rest of the rows (except the ``ordinary'' row) in Table~\ref{tab:gsac2_compare}, we notice that STA has three times lower failure rates compared to SAC in most of the settings. In addition, since we only sample $N_c=5$ candidates, the training time is almost the same as SAC. Hence, we claim that we achieve a better relay-generalization than SAC without losing the ordinary performance and the converging speed. We also compare the results between STA and the naive method. We find out that our method has a better performance in harder environments. For example, STA has only $25.2\%$ failure rate on the Humanoid SA-PPO.
However, in the easiest environment, Hopper, the naive method with $N_a=16$ pretrained agents has a twice lower rate than STA. We suggest it is because Hopper has the simplest robot that only has one lag. Hence, we suggest that most of the hoping styles have a similar version in the 16 pretrained agents. Therefore, the naive method can easily play on other agents' trajectories.

\begin{table}[t]
\caption{Failure rates (\%) of STA with different $N_c$, where $N_c$ is the number of starting candidate states we consider before starting a trajectory. State with the lowest $Q(s, \pi(s))$ is chosen among the $N_c$ candidates to start the agent. A larger $N_c$ produces better performance with negligible cost.}
\vspace{-0.4cm}

\label{tab:gsac2_sn}
\begin{center}
\resizebox{.83\textwidth}{!}{

\begin{tabular}{ccccccc}
\multicolumn{1}{c}{\multirow{2}{*}{Environment}} &
\multicolumn{1}{c}{\multirow{2}{*}{\shortstack{Stranger\\Algorithm}}}  & \multicolumn{5}{c}{Test Algorithm (Failure Rate \%)} \\
\cmidrule{3-7}
  &   & \multicolumn{1}{c}{SAC} & \multicolumn{1}{c}{$\text{Naive}_{N_a=16}$} & \multicolumn{1}{c}{$\text{STA}_{N_c=1}$} & \multicolumn{1}{c}{$\text{STA}_{N_c=2}$} & \multicolumn{1}{c}{$\text{STA}_{N_c=5}$} \\
\midrule
\multirow{4}*{\shortstack[c]{Humanoid}} 
& SAC    & 38.0 $\pm$ 33.9 & 13.2 $\pm$ 6.84 &  19.7 $\pm$ 18.1 & 8.52 $\pm$ 5.74 & \textbf{5.77 $\pm$ 3.06}    \\
& TD3    & 33.6 $\pm$ 28.5 & 7.81 $\pm$ 4.76 &  9.29 $\pm$ 7.94 & \textbf{3.48 $\pm$ 2.89} & 3.76 $\pm$ 4.49     \\
& PPO    & 48.8 $\pm$ 30.1 & 25.0 $\pm$ 3.36 &  26.1 $\pm$ 10.6 & 15.7 $\pm$ 5.25 & \textbf{9.95 $\pm$ 5.38}    \\
& SA-PPO & 83.8 $\pm$ 18.0 & 58.1 $\pm$ 8.32 &  64.2 $\pm$ 8.57 & 48.5 $\pm$ 7.44 & \textbf{25.2 $\pm$ 6.47}  \\
\midrule[0.1pt]
\multirow{4}*{\shortstack[c]{Walker2d}} 
& SAC    & 26.9 $\pm$ 23.5 & 5.38 $\pm$ 3.58 & 20.1 $\pm$ 12.2 & 16.6 $\pm$ 9.47 & \textbf{9.62 $\pm$ 3.72}  \\
& TD3    & 22.7 $\pm$ 21.7 & 3.71 $\pm$ 2.48 & 16.3 $\pm$ 10.3 & 13.1 $\pm$ 6.16 & \textbf{10.0 $\pm$ 5.81}   \\
& PPO    & 14.5 $\pm$ 12.9 & 1.19 $\pm$ 0.86 & 12.1 $\pm$ 9.35 & 8.95 $\pm$ 6.09 & \textbf{5.90 $\pm$ 2.64}   \\
& ATLA-PPO & 19.9 $\pm$ 18.1 & 2.33 $\pm$ 0.99 & 15.8 $\pm$ 9.78 & 10.4 $\pm$ 7.28 & \textbf{7.67 $\pm$ 3.90}  \\
\midrule[0.1pt]
\multirow{4}*{\shortstack[c]{Hopper}} 
& SAC    & 32.6 $\pm$ 36.0 & 2.59 $\pm$ 2.41 &  \textbf{13.61 $\pm$ 10.2} & 19.8 $\pm$ 16.0 & 14.1 $\pm$ 8.12   \\
& TD3    & 31.9 $\pm$ 36.5 & 3.05 $\pm$ 5.03 &  22.1 $\pm$ 21.0 & 20.9 $\pm$ 17.3 & \textbf{17.6 $\pm$ 12.3}  \\
& PPO    & 23.8 $\pm$ 33.5 & 2.62 $\pm$ 2.95 &  \textbf{12.6 $\pm$ 14.3} & 18.1 $\pm$ 18.1 & 12.8 $\pm$ 12.5   \\
& ATLA-PPO & 30.6 $\pm$ 35.0 & 4.69 $\pm$ 4.01 &  24.2 $\pm$ 20.5 & 23.5 $\pm$ 21.8 & \textbf{21.2 $\pm$ 16.0}   \\
\midrule
\end{tabular}
}
\end{center}
\end{table}

Besides showing the best performance of STA, we also conduct some ablation studies. First, we directly observe how STA takes over the control of other agents. Same as SAC agents, we notice that STA agents also run with one kind of posture. Even when the STA starts from other agents' posture, it will change it into its own and keep running. We consider this a good thing.  For example, if a test agent is faster than the stranger agent, we will want the test agent to run in its own way, which is faster. We also conduct an experiment on different numbers of candidates $N_c$ to show the importance of starting the trajectories from the states that has a lower $Q(s, \pi(s))$. The results are shown in Table~\ref{tab:gsac2_sn}. We first look at the column where $N_c = 1$. It shows the results of simply randomly selecting a state from the initial set as we did in the naive method. Compared to SAC, $N_c = 1$ is only slightly better. However, as we use a larger $N_c$, the result becomes comparable to the naive method, especially on Humanoid. We also notice that different $N_c$ have little effect on Hopper. This is because there is only a small $Q(s, \pi(s))$ difference between the familiar states and the unfamiliar state according to Table~\ref{tab:detect_results}. Hence, we should consider other methods in Hopper.

\section{Related Work}
\vspace{-1em}
Many methods have been designed to evaluate the generalization of an RL agent \citep{Kirk2021ASO}.
They can be categorized by which part of the RL environment are different between training and testing, including (1) the environment, (2) the observation emission function, (3) the transition function, (4) the reward function, and (5) the initial state set. For \textbf{(1)}, researchers studied training and testing in totally different environments (\citet{Cobbe2019QuantifyingGI}, and \citet{Dennis2020EmergentCA}).
For \textbf{(2)}, researchers change the surface of the objects in the environment~\citep{Tobin2017DomainRF, Zhang2018NaturalEB, Zhao2019InvestigatingGI, James2019SimToRealVS}.
For \textbf{(3)}, researchers change the parameter of the transition function like the gravity strength or the weight of the ball, or add an external force to the agent.
\citep{Packer2018AssessingGI,Pinto2017RobustAR} 
For the \textbf{(4)}, researchers~\citep{Mishra2018ASN, Dosovitskiy2017CARLAAO, ChevalierBoisvert2019BabyAIAP, Lynch2021LanguageCI} normally want to see if the agent can adapt quickly from one task to another with the general knowledge learned in the first task. 
Finally, the last category \textbf{(5)}, which only selects different initial states in the same environment~\citep{Portelas2019TeacherAF, Peng2018SimtoRealTO, Kaplanis2018ContinualRL} is where our relay-generalization belongs. In one of the most related previous work,~\citet{Zhang2018ADO} only train on limited states in the initial set $d_0$ and test on the other states in the initial set $d_0$. Their results show that if a PPO agent only trains on less than 5 states, it will be hard to avoid failing on other states in the initial set $d_0$. 
In addition, our relay-generalization belongs to Out-of-Distribution (OOD) generalization, where training and testing states are sampled from different distributions (See Figure \ref{fig:distribution_1} and Figure \ref{fig:STA_distribution}). 
To the best of our knowledge, the notion of controllable states and relay evaluation has not been discussed in the literature. 

Beyond the generalization of RL, many other RL domains are related to our works. First, in hierarchical reinforcement learning, \citet{pateria2021hierarchical, Brochu2010ATO, Nachum2018DataEfficientHR} train multiple agents with different small missions and let them complete a task together by taking turns. However, in such a scenario, the agents are trained together. Therefore, they are familiar with previous agents' trajectories. 
Second, both our naive method and the offline RL~\citep{Levine2020OfflineRL,Kumar2020ConservativeQF, ostrovski2021difficulty} leverage the trajectories of other agents.
Third, population-based training  ~\citep{jaderberg2017population, derek2021adaptable}
can be a potential way to generate controllable states and increase relay-generalization. Finally, the adversarial robustness of agents has been studied in the context of state, observation, and reward perturbations~\citep{huang2017adversarial,ilahi2021challenges,Zhang2020RobustDR,Zhang2021RobustRL,ding2022robust,lan2022alphazero}. However, these studies focus on the performance of agents under imperceptible adversarial perturbations. On the other hand, the states of other stranger agents can be very different from the agent's trajectory distribution. In addition, our work shows that robust agents (SA/ATLA PPO) are still bad at generalization, which is similar to some of the previous results~\citep{Korkmaz2021AdversarialTB}. 

Revisiting observed states strategy (or local access protocol~\citep{Yin2021EfficientLP}) is used in our STA algorithm. This strategy has been used in many RL algorithms. \cite{Ecoffet2019GoExploreAN} surpasses the state-of-the-art on on hard-exploration games by revisiting promising states in history. 
\cite{Tavakoli2020ExploringRD} shows that revisiting states on high return trajectories can improve PPO on a sparse-reward task.
After our publication, 
\cite{Yin2023SampleED} proposed revisiting the historical states that the agent is uncertain. 
Different from these works, our STA algorithm uses the Q function to select states that the agent used to be good at and now is not confident with.

\section{Conclusion}
We propose relay-evaluation, a proxy to evaluate RL agents' generalization performance on controllable states in a fixed environment. Relay-evaluation involves running an agent from the middle states of other independently-trained agents' high-reward trajectories. Extensive studies under the MuJoCo environment demonstrate that many representative RL algorithms have unexpectedly high failure rates under relay-evaluation, indicating a significant limitation of existing RL methods. To overcome this challenge, we propose a novel method called Self-Trajectory Augmentation (STA), which improves the generalization of existing RL agents without impacting ordinary performance. 
This paper opens up a new research direction in RL, and there are several limitations to be addressed in the future. For instance,  how to find unexplored controllable states more effectively and how to ``certify'' that an agent can work (or at least won't fail) on all controllable states are interesting future challenges. 
More importantly, how to develop better algorithms to improve relay generalization is still an open problem. 

\section*{Acknowledgement}
This work is supported in part by NSF under IIS-2008173, IIS-2048280 and an Okawa research grant.

\bibliography{iclr2023_conference}
\bibliographystyle{iclr2023_conference}
\newpage
\appendix
\section{Details for Experimental Setup}
For SAC and TD3, we follow the implementation of OpenAI Spinning Up \citep{SpinningUp2018}, which uses networks of size (256, 256) with relu units and 3 million environment interactions.  
For PPO, we follow the implementation of \citealp{engstrom2019implementation} since its PPO performance is higher than  OpenAI Spinning Up. The total number of environmental interactions is one million. 
For SA-PPO and ATLA-PPO, we follow the implementation in \citep{Zhang2021RobustRL}. SA-PPO and ATLA-PPO are representative algorithms in terms of observation robustness. That is, the agent can function normally even if a small adversarial perturbation is added to the observation. We include this kind of robust agent to show that our problem is different from the normal adversarial robustness problem in RL. For the humanoid environment, the SA-PPO algorithm we used is the SGLD algorithm. For other environments, we use ATLA-PPO.

For the \textbf{naive method}, both the pretrained agents and the agent follow the implementation of SAC of OpenAI Spinning Up. Hence, training a naive method with $N_a$ agents requires $N_a + 1$ times of training time. The default parameter settings are: $\eta_{\text{navie}} = 0.75$ since we want to include harder states; $p_0 \approx 0.9$, $L_r = 100$ since we want to sample more initial states form $\mathcal{S}^{\eta_{\text{naive}}}_{\text{pretrain}}$. 

For the \textbf{STA method}, we also follow the implementation of SAC of OpenAI Spinning Up. Since the candidate number $N_c$ we use is less than 10, the training time is almost the same as training a SAC agent. The default parameter settings are: $\eta_{\text{STA}} = 0.75$ since we want to include harder states as we did in the naive method; $p_0 \approx 0.9$, $L_r = 100$ since we want to sample more initial states form $\mathcal{S}_{STA}$; $\lambda=50 < L_r$ since we want to include the states in the trajectories that starts from states in $\mathcal{S}_{STA}$. For the STA parameters used in Table~\ref{tab:gsac2_compare}, we set the number of candidates $N_c=5$; we set the qualifying ratio $\gamma=1.0$ in Humanoid and $\gamma=1.6$ in Walker2d and Hopper.

\section{Experiments of our naive method with different playing lengths}
In this section, we show the results of our naive method with different limitations of the extra steps of the trajectories starting from starting set. 

The results are shown in Table~\ref{tab:gsac_cs}. According to the table, the failure rate will increase when $L_r$ is too small or too large. Since $L_r=100$ has the lowest failure rate in almost all settings, we will use $L_r=100$ as the default in other experiments.
\begin{table}[ht]
\caption{The results of the naive method with different time step limitation $L_r$. }
\vspace{-0.4cm}

\label{tab:gsac_cs}
\begin{center}
\resizebox{.76 \textwidth}{!}{

\begin{tabular}{ccccccc}
\multicolumn{1}{c}{\multirow{2}{*}{Environment}} &
\multicolumn{1}{c}{\multirow{2}{*}{\shortstack{Stranger\\Algorithm}}}   & \multicolumn{5}{c}{Test Algorithm (Failure Rate \%)} \\
\cmidrule{3-7}
   &   & \multicolumn{1}{c}{SAC} & \multicolumn{1}{c}{$\text{Naive}_{L_r=50}$} & \multicolumn{1}{c}{$\text{Naive}_{L_r=100}$} & \multicolumn{1}{c}{$\text{Naive}_{L_r=200}$} & \multicolumn{1}{c}{$\text{Naive}_{L_r=500}$} \\
\midrule
\multirow{4}*{\shortstack[c]{Humanoid}} 
& SAC    & 38.0 $\pm$ 33.9 & 30.4 $\pm$ 34.9 & \textbf{13.2 $\pm$ 6.84} & 14.3 $\pm$ 6.00 & 14.76 $\pm$ 5.08   \\
& TD3    & 33.6 $\pm$ 28.5 & 25.7 $\pm$ 37.3 & \textbf{7.81 $\pm$ 4.76} & 10.6 $\pm$ 3.74 & 11.76 $\pm$ 6.61    \\
& PPO    & 48.8 $\pm$ 30.1 & 40.6 $\pm$ 30.0 & \textbf{25.0 $\pm$ 3.36} & 25.0 $\pm$ 5.04 & 28.36 $\pm$ 7.31   \\
& SA-PPO & 83.8 $\pm$ 18.0 & 69.7 $\pm$ 15.7 & \textbf{58.1 $\pm$ 8.32} & 61.8 $\pm$ 12.2 & 65.43 $\pm$ 10.8 \\
\midrule[0.1pt]
\multirow{4}*{\shortstack[c]{Walker2d}} 
& SAC    & 26.9 $\pm$ 23.5 & 6.10 $\pm$ 2.34 & 5.38 $\pm$ 3.58 & \textbf{4.48 $\pm$ 3.15} & 6.52 $\pm$ 2.87  \\
& TD3    & 22.7 $\pm$ 21.7 & 3.43 $\pm$ 1.52 & 3.71 $\pm$ 2.48 & \textbf{3.14 $\pm$ 1.80} & 3.48 $\pm$ 2.20  \\
& PPO    & 14.5 $\pm$ 12.9 & 1.95 $\pm$ 1.35 & \textbf{1.19 $\pm$ 0.86} & 1.81 $\pm$ 1.06 & 2.81 $\pm$ 1.97  \\
& ATLA & 19.9 $\pm$ 18.1 & 3.36 $\pm$ 1.94 & \textbf{2.33 $\pm$ 0.99} & 2.88 $\pm$ 1.24 & 3.78 $\pm$ 2.24  \\
\midrule[0.1pt]
\multirow{4}*{\shortstack[c]{Hopper}} 
& SAC    & 32.6 $\pm$ 36.0 & 4.35 $\pm$ 4.73 & \textbf{2.59 $\pm$ 2.41} & 7.96 $\pm$ 5.68 & 7.82 $\pm$ 6.54 \\
& TD3    & 31.9 $\pm$ 36.5 & 3.76 $\pm$ 5.74 & \textbf{3.05 $\pm$ 5.03} & 5.19 $\pm$ 4.37 & 6.52 $\pm$ 9.11 \\
& PPO    & 23.8 $\pm$ 33.5 & 3.14 $\pm$ 4.57 & \textbf{2.62 $\pm$ 2.95} & 5.86 $\pm$ 5.15 & 3.71 $\pm$ 6.79 \\
& ATLA & 30.6 $\pm$ 35.0 & \textbf{3.67 $\pm$ 3.31} & 4.69 $\pm$ 4.01 & 6.05 $\pm$ 3.88 & 5.85 $\pm$ 7.50 \\
\midrule
\end{tabular}
}
\end{center}
\end{table}

\newpage
\section{Experiments of STA with Different Candidate Criteria}
In this section, we evaluate the performance of using different $\gamma$. In STA, we use $\omega_{max}/\gamma$ to serve as the criteria to be the threshold of being one of the $N_c$ candidates. Only states with score $\omega_{max}/\gamma$ in $\mathcal{S}_{\text{STA}}$ can be the candidate of the initial state $s_0$. If  $\gamma$ is larger, then states with lower scores can also be the candidates.

The results different $\gamma$ with $N_c = 5$ are shown in Table~\ref{tab:gsac2_owr}. According to the results, the best $\gamma$ of each environment is different. For example, in Humanoid, it is important to only samples the states that have higher scores as candidates (small $\gamma$).

\begin{table}[h]
\caption{The results of STA with different $\gamma$. When $\gamma$ is larger, the states with lower scores can also
be the candidates of being the initial state of a trajectory during training STA. }
\vspace{-0.4cm}

\label{tab:gsac2_owr}
\begin{center}
\resizebox{.85\textwidth}{!}{

\begin{tabular}{cccccccc}
\multicolumn{1}{c}{\multirow{2}{*}{Environment}} &
\multicolumn{1}{c}{\multirow{2}{*}{\shortstack{Stranger\\Algorithm}}} & \multicolumn{6}{c}{Test Algorithm (Failure Rate \%)} \\
\cmidrule{3-8}
   &  & \multicolumn{1}{c}{SAC} & \multicolumn{1}{c}{Naive} & \multicolumn{1}{c}{$\gamma = 1.0$}  & \multicolumn{1}{c}{$\gamma = 1.1$} & \multicolumn{1}{c}{$\gamma = 1.3$} & \multicolumn{1}{c}{$\gamma = 1.6$} \\
\midrule
\multirow{4}*{\shortstack[c]{Humanoid}} 
& SAC    & 38.0 $\pm$ 33.9 & 13.2 $\pm$ 6.84 &  5.77 $\pm$ 3.06 & \textbf{4.44 $\pm$ 5.34} & 8.36 $\pm$ 5.62 & 9.10 $\pm$ 5.55    \\
& TD3    & 33.6 $\pm$ 28.5 & 7.81 $\pm$ 4.76 &  3.76 $\pm$ 4.49 & \textbf{2.24 $\pm$ 2.67} & 6.05 $\pm$ 4.30 & 3.62 $\pm$ 3.56   \\
& PPO    & 48.8 $\pm$ 30.1 & 25.0 $\pm$ 3.36 &  \textbf{9.95 $\pm$ 5.38} & 12.0 $\pm$ 6.33 & 15.8 $\pm$ 5.20 & 12.4 $\pm$ 5.61     \\
& SA-PPO & 83.8 $\pm$ 18.0 & 58.1 $\pm$ 8.32 &  \textbf{25.2 $\pm$ 6.47} & 33.2 $\pm$ 15.2 & 48.5 $\pm$ 11.4 & 43.6 $\pm$ 6.97   \\
\midrule[0.1pt]
\multirow{4}*{\shortstack[c]{Walker2d}} 
& SAC    & 26.9 $\pm$ 23.5 & 5.38 $\pm$ 3.58 & 9.62 $\pm$ 3.72 & 11.5 $\pm$ 6.92 & \textbf{6.52 $\pm$ 3.92} & 7.33 $\pm$ 4.62   \\
& TD3    & 22.7 $\pm$ 21.7 & 3.71 $\pm$ 2.48 & 10.0 $\pm$ 5.81 & 11.5 $\pm$ 7.70 & 6.38 $\pm$ 5.76 & \textbf{5.48 $\pm$ 3.35}    \\
& PPO    & 14.5 $\pm$ 12.9 & 1.19 $\pm$ 0.86 & 5.90 $\pm$ 2.64 & 7.00 $\pm$ 4.10 & 5.24 $\pm$ 4.69 & \textbf{2.57 $\pm$ 3.25} \\
& ATLA & 19.9 $\pm$ 18.1 & 2.33 $\pm$ 0.99 & 7.67 $\pm$ 3.90 & 8.17 $\pm$ 5.16 & 7.49 $\pm$ 5.27 & \textbf{2.83 $\pm$ 2.04} \\
\midrule[0.1pt]
\multirow{4}*{\shortstack[c]{Hopper}} 
& SAC    & 32.6 $\pm$ 36.0 & 2.59 $\pm$ 2.41 &  14.1 $\pm$ 8.12 & 23.6 $\pm$ 15.1 & 23.2 $\pm$ 10.6 & \textbf{9.80 $\pm$ 4.51}  \\
& TD3    & 31.9 $\pm$ 36.5 & 3.05 $\pm$ 5.03 &  17.6 $\pm$ 12.3 & 18.3 $\pm$ 19.2 & 28.5 $\pm$ 19.5 & \textbf{13.8 $\pm$ 8.20}   \\
& PPO    & 23.8 $\pm$ 33.5 & 2.62 $\pm$ 2.95 &  12.8 $\pm$ 12.5 & 20.2 $\pm$ 16.3 & 21.2 $\pm$ 12.6 & \textbf{8.52 $\pm$ 6.04}   \\
& ATLA & 30.6 $\pm$ 35.0 & 4.69 $\pm$ 4.01 &  21.2 $\pm$ 16.0 & 21.0 $\pm$ 17.2 & 28.9 $\pm$ 18.7 & \textbf{10.2 $\pm$ 9.48}   \\
\midrule
\end{tabular}
}
\end{center}
\end{table}

\section{Infinity replay buffer}
In this section, we show that our STA method is better than using an unlimited-size of replay buffer. The original replay buffer size is one million. The maximum size of an unlimited-size of replay buffer is three million since we only have three million environment interactions. The results are shown in Table~\ref{tab:sac_inf}. According to the results, SAC with an infinite replay buffer still has poor performance on relay evaluation compared to our method. Moreover, the average return under ordinary evaluation is lower than the normal SAC, while our method is greater or equal to the normal SAC.

\begin{table}[h]
\caption{SAC with infinity replay buffer during training.}
\vspace{-0.4cm}

\label{tab:sac_inf}
\begin{center}
\resizebox{.77\textwidth}{!}{

\begin{tabular}{cccccc}
\multicolumn{1}{c}{\multirow{2}{*}{Environment}} &
\multicolumn{1}{c}{\multirow{2}{*}{\shortstack{Stranger\\Algorithm}}}  & \multicolumn{4}{c}{Test Algorithm (Failure Rate \%)} \\
\cmidrule{3-6}
  &   & \multicolumn{1}{c}{SAC} & \multicolumn{1}{c}{$\text{Naive}_{N_a=16}$} & \multicolumn{1}{c}{$\text{STA}$} & \multicolumn{1}{c}{$\text{SAC}_{\text{replay\_buffer}=\infty}$}  \\
\midrule
\multirow{4}*{\shortstack[c]{Humanoid}} 
& SAC    & 38.0 $\pm$ 33.9 & 13.2 $\pm$ 6.84 & 5.77   $\pm$ 23.3 & 34.07 $\pm$ 10.12    \\
& TD3    & 33.6 $\pm$ 28.5 & 7.81 $\pm$ 4.76 & 3.76   $\pm$ 19.0 & 24.00 $\pm$ 15.67     \\
& PPO    & 48.8 $\pm$ 30.1 & 25.0 $\pm$ 3.36 & 9.95   $\pm$ 29.9 & 43.39 $\pm$ 12.31    \\
& SA-PPO & 83.8 $\pm$ 18.0 & 58.1 $\pm$ 8.32 & 25.2   $\pm$ 43.4 & 80.10 $\pm$ 8.43   \\
\midrule[0.1pt]
\multirow{4}*{\shortstack[c]{Walker2d}} 
& SAC    & 26.9 $\pm$ 23.5 & 5.38 $\pm$ 3.58   &  7.33 $\pm$ 26.0 & 41.33 $\pm$ 26.87    \\
& TD3    & 22.7 $\pm$ 21.7 & 3.71 $\pm$ 2.48   &  5.48 $\pm$ 22.7 & 38.19 $\pm$ 27.21    \\
& PPO    & 14.5 $\pm$ 12.9 & 1.19 $\pm$ 0.86   &  2.57 $\pm$ 15.8 & 24.57 $\pm$ 23.94    \\
& ATLA-PPO & 19.9 $\pm$ 18.1 & 2.33 $\pm$ 0.99 &  2.83 $\pm$ 16.5 & 28.47 $\pm$ 25.46    \\
\midrule
\end{tabular}
}
\end{center}
\end{table}

\newpage
\section{Using Average Reward as Scoring Function}
In this section, we show the STA result of using the average score $\text{score}(s_t) = \sum_{i=t}^{T-1} r_i / (T-t)$ instead of using the reward sum of next $\lambda$ steps $\text{score}(s_t) = \sum_{i=t}^{t+\lambda-1} r_i$. In STA, the scoring function is a quick way to estimate if a state is controllable. We only add the states that have high enough scores into the $\mathcal{S}_{\text{STA}}$. 
According to the results, using the average reward can also reduce the failure rate compared to the original SAC. However, its failure rate is still higher than using our score function (column STA), which uses the sum of the reward of the next $\lambda$ step.

\begin{table}[h]
\caption{The experiment of STA that use average rewards $\text{score}(s_t) = \sum_{i=t}^{T-1} r_i / (T-t)$ as the score function to estimate if a state in $\mathcal{S}_{\text{STA}}$ is controllable.}
\vspace{-0.4cm}

\label{tab:sta_avg_ret_score}
\begin{center}
\resizebox{.77\textwidth}{!}{

\begin{tabular}{cccccc}
\multicolumn{1}{c}{\multirow{2}{*}{Environment}} &
\multicolumn{1}{c}{\multirow{2}{*}{\shortstack{Stranger\\Algorithm}}}  & \multicolumn{4}{c}{Test Algorithm (Failure Rate \%)} \\
\cmidrule{3-6}
  &   & \multicolumn{1}{c}{SAC} & \multicolumn{1}{c}{$\text{Naive}_{N_a=16}$} & \multicolumn{1}{c}{$\text{STA}$} & \multicolumn{1}{c}{$\text{STA}_{\text{average reward}}$}  \\
\midrule
\multirow{4}*{\shortstack[c]{Humanoid}} 
& SAC    & 38.0 $\pm$ 33.9 & 13.2 $\pm$ 6.84 & 5.77   $\pm$ 23.3 & 8.36 $\pm$ 3.52    \\
& TD3    & 33.6 $\pm$ 28.5 & 7.81 $\pm$ 4.76 & 3.76   $\pm$ 19.0 & 4.86 $\pm$ 2.30    \\
& PPO    & 48.8 $\pm$ 30.1 & 25.0 $\pm$ 3.36 & 9.95   $\pm$ 29.9 &  11.85 $\pm$ 4.72    \\
& SA-PPO & 83.8 $\pm$ 18.0 & 58.1 $\pm$ 8.32 & 25.2   $\pm$ 43.4 &  44.29 $\pm$ 11.14  \\
\midrule[0.1pt]
\end{tabular}
}
\end{center}
\end{table}

\section{The failure definition of Mujoco}
In Mujoco, by default, most of the environments will terminate the simulation when the agent is unhealthy. Normally, unhealthy means the robot has fallen over. By doing so, the agent can start a new trajectory without wasting more time on the "unhealthy" trajectory. Each environment has its own way of detecting unhealthy. 

The Humanoid environment requires the z-position (height) of the robot need to be in a predefined range.

The Ant and Walker2d environment defines the robot as unhealthy if any of the following happens:
\begin{itemize}
  \item Any of the state space values is no longer finite.
  \item The height of the walker is not in a predefined range.
  \item The absolute value of the angle is not in a predefined range. 
\end{itemize}

The Hopper environment defines the robot as unhealthy if any of the following happens:
\begin{itemize}
  \item An element of observation[1:] is not in a predefined range.
  \item The height of the walker is not in a predefined range.
  \item The absolute value of the angle is not in a predefined range. 
\end{itemize}

Please see more details at https://www.gymlibrary.dev/environments/mujoco/.

\newpage
\section{Define Failure with Returns}
In this section, we define the failure of relay-evaluation according to the reward sum of the next $L=500$ steps instead of using the failure defined by Mujoco.
The first column shows the failure rate using the failure definition of the Mujoco. For the remaining columns, we allow the agent to keep playing even if the current state is "failed," according to Mujoco. 
In the end, we define a trajectory as a failure trajectory according to its return. For example, for the second column, we define a trajectory as failed when its return is lower than 500. Note that the average return of SAC should be 2828. 
According to the results, no matter using which kind of definition of failure, SAC still has an unacceptable failure rate, especially on the SA-PPO. This shows that the termination criteria in the MuJoCo environments are not too strict and most of the terminated trajectories have no chance to get a high score.

\begin{table}[h]
\caption{Conducting relay-evaluation with different definitions of failure. The column SAC$_{\text{Mujoco}}$ uses the failure defined by the environment. The columns SAC$_{<\text{thr}}$ define a trajectory is failed if the return value is smaller than the threshold thr. Note that the expected return value is greater than 2800.  
}
\vspace{-0.4cm}

\label{tab:different_failure_define}
\begin{center}
\resizebox{.77\textwidth}{!}{

\begin{tabular}{cccccc}
\multicolumn{1}{c}{\multirow{2}{*}{Environment}} &
\multicolumn{1}{c}{\multirow{2}{*}{\shortstack{Stranger\\Algorithm}}}  & \multicolumn{4}{c}{Test Algorithm (Failure Rate \%)} \\
\cmidrule{3-6}
  &   & \multicolumn{1}{c}{SAC$_{\text{Mujoco}}$} & \multicolumn{1}{c}{SAC$_{<500}$} & \multicolumn{1}{c}{SAC$_{<1000}$} & \multicolumn{1}{c}{SAC$_{<1500}$}  \\
\midrule
\multirow{4}*{\shortstack[c]{Humanoid}} 
& SAC    & 38.0 $\pm$ 33.9 & 30.56 $\pm$ 11.11 &   32.28 $\pm$ 11.12 &  32.54 $\pm$ 10.95   \\
& TD3    & 33.6 $\pm$ 28.5 & 21.55 $\pm$ 5.77  &   23.93 $\pm$ 5.67  &  24.09 $\pm$ 5.76  \\
& PPO    & 48.8 $\pm$ 30.1 & 42.42 $\pm$ 10.02 &   44.93 $\pm$ 10.41 &  45.33 $\pm$ 10.48  \\
& SA-PPO & 83.8 $\pm$ 18.0 & 81.83 $\pm$ 9.29  &   82.82 $\pm$ 9.03  &  82.98 $\pm$ 9.04 \\
\midrule[0.1pt]
\end{tabular}
}
\end{center}
\end{table}

\section{State Distribution of an STA Agent's Old Trajectories}
In this section, we plot the historical states of an STA agent with the states we showed in Fig~\ref{fig:distribution_1}. As the figure shows, although the distribution of historical states of the STA agent is more diverse than the distribution of other agents' states, there are still many areas that it does not cover. Hence, this suggests that the STA agents have a better performance on relay evaluation due to being more general.  
\begin{figure}[h]
\caption{State distribution of an STA agent's $\mathcal{S}_{\text{STA}}$. $\mathcal{S}_{\text{STA}}$ stores the likely controllable states that the agent generated during training. We use t-SNE to visualize those states along with the states generated by the other six agents used in Fig.1a.
}
\label{fig:STA_distribution}
\centering
\includegraphics[width=0.6\textwidth]{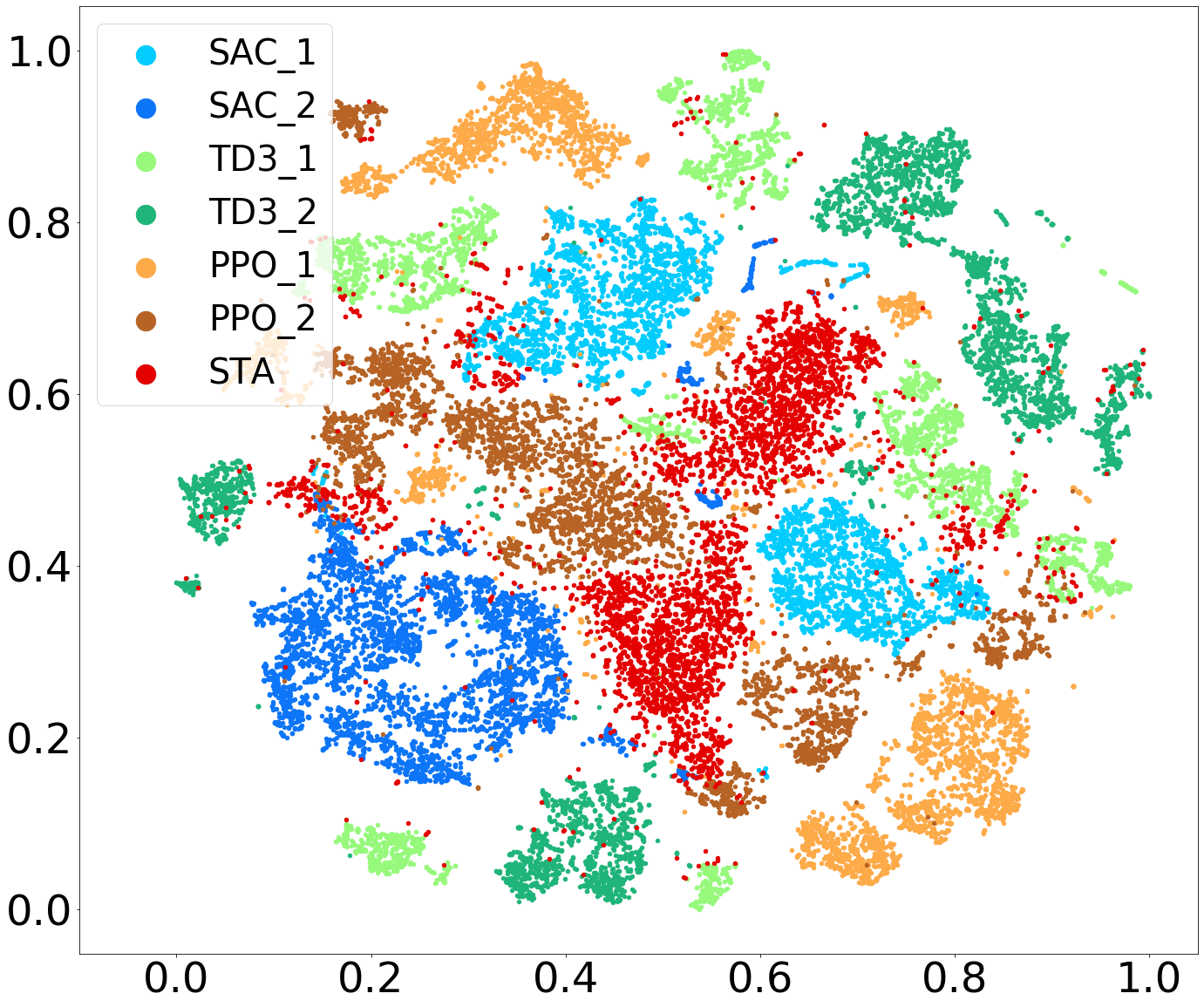}
\end{figure}
\newpage
\section{Evaluating SAC with Different L in Relay-Evaluation}
In this section, we conduct relay-evaluations with different $L$. We select SAC as our test agent. Our result is shown in Table~\ref{tab:different_L}. The results show that most of the failures happened in the first 100 steps after the test agent took over the control. The results also show that for some of the cases, even if the agent does not fail for 100 steps, it may still fail in the next 400 steps. 

\begin{table}[h]
\caption{Evaluating SAC with different L in relay-evaluation, where L is the extra steps that the test agent needs to complete without failing.}
\vspace{-0.4cm}

\label{tab:different_L}
\begin{center}
\resizebox{.77\textwidth}{!}{

\begin{tabular}{cccccc}
\multicolumn{1}{c}{\multirow{2}{*}{Environment}} &
\multicolumn{1}{c}{\multirow{2}{*}{\shortstack{Stranger\\Algorithm}}}  & \multicolumn{4}{c}{Test Algorithm (Failure Rate \%)} \\
\cmidrule{3-6}
  &   & \multicolumn{1}{c}{SAC$_{L=500}$} & \multicolumn{1}{c}{SAC$_{L=200}$} & \multicolumn{1}{c}{SAC$_{L=100}$} & \multicolumn{1}{c}{SAC$_{L=50}$}  \\
\midrule
\multirow{4}*{\shortstack[c]{Humanoid}} 
& SAC    & 38.0 $\pm$ 33.9 & 32.32 $\pm$ 11.08 &   30.69 $\pm$ 10.92  &  21.12 $\pm$ 7.25    \\
& TD3    & 33.6 $\pm$ 28.5 & 23.97 $\pm$ 5.71 &   21.71 $\pm$ 6.22  &  11.19 $\pm$ 6.09   \\
& PPO    & 48.8 $\pm$ 30.1 & 45.06 $\pm$ 10.59 &   42.50 $\pm$ 10.26  &  28.84 $\pm$ 7.81    \\
& SA-PPO & 83.8 $\pm$ 18.0 & 82.90 $\pm$ 9.03 &   82.02 $\pm$ 9.05   &  73.77 $\pm$ 9.60 \\
\midrule[0.1pt]
\end{tabular}
}
\end{center}
\end{table}

\section{Generating Controllable States with Different Hyperparameters}
In this section, our goal is to generate more controllable states to test our agents by training with different hyperparameters. We first train 40 SAC agents with random hyperparameters and use the top 10 agents to serve as stranger agents. 

The hyperparameters that we will randomly choices are:
the learning rate of the policy $\pi$, the learning rate of the Q function, the batch size, the network structure, the training frequency, and the $\alpha$ in the SAC goal function.

The results are shown in Table~\ref{tab:PBT}. According to Table 12, we observe that the states generated by random hyperparameters are easier for other agents to pass the relay-evaluation, especially when evaluating SAC and TD3 agents. A potential reason is that some of these hyperparameters lead to worse policies and generate slower trajectories, and it’s easier for the target agent to take over from those slower states. 

\begin{table}[h]
\caption{We generate controllable states by training SAC with random hyperparameters, including learning rates, network sizes, and batch size.}
\vspace{-0.4cm}

\label{tab:PBT}
\begin{center}
\resizebox{.85\textwidth}{!}{

\begin{tabular}{ccccccc}

\multicolumn{1}{c}{\multirow{2}{*}{Environment}} &
\multicolumn{1}{c}{\multirow{2}{*}{\shortstack{Stranger\\Algorithm}}} &
\multicolumn{4}{c}{Test Agent Algorithm (Failure Rate \%)} \\
\cmidrule{3-6}

& & \multicolumn{1}{c}{SAC} & \multicolumn{1}{c}{TD3} & \multicolumn{1}{c}{PPO} & \multicolumn{1}{c}{SA-PPO/ATLA} \\
\midrule
\multirow{5}*{\shortstack[c]{Humanoid}} 
& SAC      & \textbf{38.0 $\pm$ 33.9} & 83.9 $\pm$ 17.6 & 83.9 $\pm$ 16.5 & 65.1 $\pm$ 31.8 \\
& TD3      & \textbf{33.6 $\pm$ 28.5} & 60.5 $\pm$ 30.1 & 78.4 $\pm$ 20.0 & 67.5 $\pm$ 29.9 \\
& PPO       & \textbf{48.8 $\pm$ 30.1} & 77.8 $\pm$ 24.3 & 81.6 $\pm$ 19.1 & 63.2 $\pm$ 30.9  \\
& SA-PPO &  83.8 $\pm$ 18.0          & 96.2 $\pm$ 5.66 & 92.9 $\pm$ 11.9 & \textbf{77.0 $\pm$ 26.4} \\
& SAC-Random-Hyperparameters &  \textbf{16.6 $\pm$ 37.2} & 55.8 $\pm$ 49.6 & 79.5 $\pm$ 40.3 & 65.0 $\pm$ 47.6 \\
\midrule
\end{tabular}
}
\end{center}
\end{table}

\section{The ordinary return of the test agents in Table~\ref{tab:main_result}}

This section shows the average returns of the agents we used in Table~\ref{tab:main_result}.

\begin{table}[h]
\caption{The avg return of the test agents we used in Table~\ref{tab:main_result}.}
\vspace{-0.4cm}

\label{tab:test_agent_ret}
\begin{center}
\resizebox{.77\textwidth}{!}{

\begin{tabular}{cccccc}
Environment & SAC & TD3 & PPO & SA/ATLA PPO  \\ 
\midrule[0.1pt]
Humanoid    & 5645.73 $\pm$ 233.54 & 5011.30 $\pm$ 1515.59 & 5173.36 $\pm$ 236.10 & 6614.09 $\pm$ 340.69   \\
Walker2d    & 5715.64 $\pm$ 446.39 & 5588.48 $\pm$ 701.70 & 4184.67 $\pm$ 897.59 & 3636.84 $\pm$ 1140.55  \\
Hopper      & 3667.81 $\pm$ 143.62 & 3520.63 $\pm$ 177.80 & 5384.12 $\pm$ 144.40 & 4587.02 $\pm$ 271.10    \\
Ant         & 6155.81 $\pm$ 176.08 & 6007.45 $\pm$ 770.93 & 3148.41 $\pm$ 404.15 &  2721.81 $\pm$ 900.34  \\
\midrule[0.1pt]

\end{tabular}
}
\end{center}
\end{table}
\end{document}